\documentclass[10pt,twocolumn,letterpaper]{article}

\usepackage[pagenumbers]{cvpr} %

\usepackage{multirow}
\usepackage{tikz}
\usepackage{pgfplots}
\pgfplotsset{compat=1.18}
\usepackage{xcolor}
\usepackage{subcaption}
\usepackage{array}
\usepackage[accsupp]{axessibility}  %

\newcolumntype{H}{>{\setbox0=\hbox\bgroup}c<{\egroup}@{}}

\definecolor{snsBlueColor}{rgb}{0.2980392156862745, 0.4470588235294118, 0.6901960784313725}
\definecolor{snsGreenColor}{rgb}{0.3333333333333333, 0.6588235294117647, 0.40784313725490196}

\newcommand{\OURS}{CrossFlow\xspace}
\newcommand{\BASELINEShort}{Standard FM\xspace}
\newcommand{\BASELINE}{Standard Flow Matching\xspace}

\newcommand{\Sota}{State-of-the-art\xspace}
\newcommand{\sota}{state-of-the-art\xspace}
\newcommand{\ttoi}{text-to-image\xspace}
\newcommand{\ttoishort}{T2I\xspace}

\newcommand{\Dm}{Diffusion model\xspace}
\newcommand{\FM}{flow matching\xspace}  %
\newcommand{\FMCaps}{Flow Matching\xspace}
\newcommand{\fm}{flow matching\xspace}

\newcommand{\XFM}{cross-modal flow matching\xspace}  %

\newcommand{\xfm}{cross-modal flow matching\xspace}
\newcommand{\VE}{Variational Encoder\xspace}
\newcommand{\VEShort}{VE\xspace}
\newcommand{\CFG}{Classifier-free guidance\xspace}
\newcommand{\CFGShort}{CFG\xspace}
\newcommand{\llamathree}{Llama3\xspace}
\newcommand{\CLIP}{CLIP\xspace}

\definecolor{cvprblue}{rgb}{0.21,0.49,0.74}
\usepackage[pagebackref,breaklinks,colorlinks,allcolors=cvprblue]{hyperref}

\def\paperID{4452} %
\def\confName{CVPR}
\def\confYear{2025}

\title{Flowing from Words to Pixels: \\ A Noise-Free Framework for Cross-Modality Evolution}

\author{
  Qihao Liu$^{1, 2}$\quad
  Xi Yin$^1$\quad
  Alan Yuille$^2$\quad
  Andrew Brown$^1$\quad
  Mannat Singh$^1$ \\
 {\normalsize $^1$GenAI, Meta \qquad $^2$Johns Hopkins University} \\
 {\small \texttt{\url{https://cross-flow.github.io/}}}
}

\begin{document}

\twocolumn[{%
\renewcommand\twocolumn[1][]{#1}%
\maketitle
\begin{center}
    \centering
    \vspace{-6mm}
    \captionsetup{type=figure}
    \includegraphics[width=\linewidth]{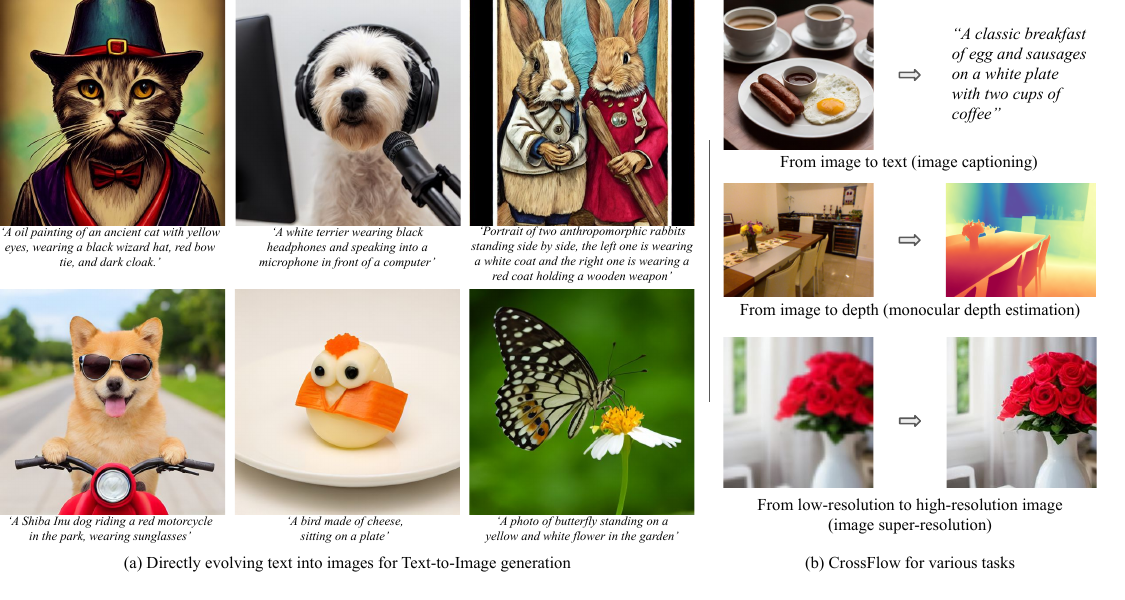}
    \vspace{-10mm}
    \captionof{figure}{\textbf{We propose \OURS, a general and simple framework that directly evolves one modality to another using \FM with no additional conditioning}.
    This is enabled using a vanilla transformer without cross-attention, achieving comparable performance with \sota models on (a) \ttoi generation, and (b) various other tasks, without requiring task specific architectures. 
    }
    \label{fig:teaser}
\end{center}%
}]

\begin{abstract}
\Dm{}s, and their generalization, \fm, have had a remarkable impact on the field of media generation. 
Here, the conventional approach is to learn the complex mapping from a simple source distribution of Gaussian noise to the target media distribution.
For cross-modal tasks such as text-to-image generation, this same mapping from noise to image is learnt whilst including a conditioning mechanism in the model. 
One key and thus far relatively unexplored feature of \fm is that, unlike \Dm{}s, they are not constrained for the source distribution to be noise.
Hence, in this paper, we propose a paradigm shift, and ask the question of whether we can instead train \fm models to learn a direct mapping from the distribution of one modality to the distribution of another, thus obviating the need for both the noise distribution and conditioning mechanism. 
We present a general and simple framework, \OURS, for \xfm. 
We show the importance of applying \VE{}s to the input data, and introduce a method to enable \CFG.
Surprisingly, for \ttoi, \OURS with a vanilla transformer without cross attention slightly outperforms standard \fm, and we show that it scales better with training steps and model size, while also allowing for interesting latent arithmetic which results in semantically meaningful edits in the output space.
To demonstrate the generalizability of our approach, we also show that \OURS is on par with or outperforms the \sota for various cross-modal / intra-modal mapping tasks, viz. image captioning, depth estimation, and image super-resolution. 
We hope this paper contributes to accelerating progress in cross-modal media generation.
\end{abstract}
    
\section{Introduction}
\label{sec:intro}

Diffusion models have achieved remarkable success in generating images~\cite{dhariwal2021diffusion, ramesh2022hierarchical, nichol2021glide, saharia2022photorealistic, rombach2022high}, videos~\cite{ho2022imagen, blattmann2023align, singer2022make, blattmann2023stable}, audio~\cite{kong2020diffwave, liu2023audioldm}, and 3D content~\cite{poole2022dreamfusion, lin2023magic3d}, revolutionizing the field of generative AI.
Recently, \FM~\cite{lipman2022flow, albergo2023building, liu2023flow} has been proposed as a generalization of diffusion models, where models are trained to find an optimal transport probability path between a source noise distribution and the target data distribution.
This approach offers simpler, straight-line trajectories compared to the complex, curved trajectories in diffusion paths. 
As a result, it has been rapidly adopted in the latest \sota image and video generation models, including LDMs~\cite{esser2024scaling} and Movie Gen~\cite{polyak2024movie}.

Both diffusion and flow-based models are typically trained to learn the mapping from noise to the target distribution.
For cross-modal generation tasks such as \ttoi~\cite{rombach2022high, chen2023pixart}, this same mapping from noise to the target modality distribution (\ie the images) is learnt whilst adding a conditioning mechanism for the conditioning modality (\ie the text) such as cross-attention. 
Unlike denoising diffusion models~\cite{ho2020denoising,song2021scorebased}, one relatively unexplored feature of \fm models is that they are not constrained for the source distribution to be Gaussian noise;
instead, the source distribution could be one that is correlated with the target distribution.
Compared to noise, learning a mapping from such a distribution should intuitively be \textit{easier} for the model because it has to learn shorter and more efficient probability paths.
A question remains however as to what this correlated source distribution could be.

Interestingly, due to the information redundancy between different modalities arising from the same data point, for cross-modal generation tasks, the provided conditioning (\eg the text in \ttoi) resembles such data that is correlated with the target distribution (\eg the images).
Hence, in this paper, we propose a paradigm shift for cross-modal generation, and ask the question of whether we can instead train \fm models to learn a direct mapping from the distribution of one modality to the distribution of another, \emph{hence obviating the need for both the noise distribution and any conditioning mechanism}.

Despite the exciting theoretical motivation, there are several key challenges in practice. 
First, both diffusion and flow-based models require the source and target distributions to be of the same shape; a requirement that is not satisfied for data distributions from different modalities.
Secondly, \sota methods heavily rely on \CFG (\CFGShort)~\cite{ho2022classifier} for improved generation quality; a method that is not compatible with \XFM due to the lack of a conditioning mechanism to turn on/off since the conditioning information instead lies \emph{within} the source data.
As a result, prior work~\cite{heitz2023iterative, liu2023flow, albergo2023building} targets the simple setting of mapping between two similar intra-modal distributions, such as human faces to cat faces~\cite{liu2023flow}.

In this work, we present key architecture design solutions for overcoming these challenges:
First, we employ a \VE for encoding the source modality data distribution to the same shape as the target modality, and show that the resulting regularization in the source distribution is essential for generation performance.
Secondly, we enable \CFGShort in \XFM through the introduction of a binary conditioning indicator during training, and demonstrate the quantitative benefits of this approach compared to alternative \CFGShort methods.
We present \OURS; a general framework for mapping between two different modalities without the need for any conditioning mechanism or noise distribution. 
Typically, different cross-modal generation tasks require task-specific architectural and training modifications, but \OURS works for different tasks without any such changes.

Using the ubiquitous albeit challenging \ttoi (\ttoishort) generation task as our primary setting, we show the significant result that \OURS outperforms commonly used \FM baselines, given the same training data, model size, and training budget, all \emph{without requiring any cross-attention layers}.
\OURS exhibits improved scaling behavior over standard \FM using cross-attention when scaling training steps or model size, and is also compatible with a variety of Large Language Models (LLMs), including CLIP~\cite{radford2021learning}, T5~\cite{raffel2020exploring}, and \llamathree~\cite{dubey2024llama}. 
Additionally, we demonstrate that since our approach encodes the source distribution into a regularized continuous space with semantic structure, \OURS enables exciting new \emph{latent arithmetic} for the \ttoi task, \eg, $\mathcal{L}$(``A dog with a hat'') + $\mathcal{L}$(``Sunglasses'') -- $\mathcal{L}$(``A hat'') creates an image of a dog wearing sunglasses without a hat. 
Lastly, \OURS\ enables bi-directional mapping between modalities, allowing, for instance, the inversion of text-to-image models to serve as image-to-text (captioning) models.

We demonstrate the general-purpose nature of \OURS on various cross-modal/intra-modal tasks:
image-to-text (image captioning), image-to-depth (depth estimation), and low-resolution to high-resolution image (super-resolution). 
\OURS achieves comparable or superior performance to various \sota methods on all three tasks, without requiring task specific architectures.
For example, in image captioning, \OURS~directly projects images into a textual latent space to generate captions, achieving \sota performance using only a simple text decoder that maps textual latents to discrete tokens.
Results are shown in~\cref{fig:teaser}.
We hope this paper contributes to accelerating the progress in cross-modal media generation.

\section{Related Work}
\label{sec:relatedWork}

\noindent\textbf{Diffusion models and rectified flow.}
Starting from Gaussian noise, diffusion~\cite{sohl2015deep, ho2020denoising} and score-based~\cite{hyvarinen2005estimation, song2019generative} generative models progressively approximate the reverse ODE of a stochastic forward process to generate data.
These models have driven significant advances across various domains, particularly in high-fidelity image~\cite{dhariwal2021diffusion, ho2022cascaded, peebles2023scalable, bao2023all, liu2024alleviating}, video~\cite{ho2022imagen, blattmann2023align, singer2022make, blattmann2023stable, polyak2024movie}, and 3D generation~\cite{poole2022dreamfusion, lin2023magic3d, liu2023one, liu2024direct}.
Recently, rectified flow models~\cite{lipman2022flow, albergo2023building, liu2023flow}, such as \FM, have been proposed to improve the generative process by enabling a transport map between two distributions.
They enable faster training and sampling by avoiding complex probability flow ODEs.

\noindent\textbf{Directly bridging distributions.} 
\FMCaps theoretically allows for arbitrary distributions as the source distribution, which can then be used for direct evolution. %
Various approaches have been proposed in this direction, such as InterFlow~\cite{albergo2023building}, $\alpha$-blending~\cite{heitz2023iterative}, data-dependent coupling~\cite{albergo2023stochastic}, and Schrödinger Bridge~\cite{liu20232, shi2024diffusion, zhou2023denoising, liu2023generalized, de2023augmented, tang2024simplified, tong2023simulation}.
They provide important theoretical support for using ODE-based methods to bridge two arbitrary distributions.
However, they are still limited to similar distributions from the same domain, such as image-to-image translation (\eg, faces-to-faces~\cite{liu2023flow,zhou2023denoising} or sketches-to-images~\cite{liu20232}). 
As a step forward, \OURS focuses on learning the mapping between data distributions arising from even different modalities.

\noindent\textbf{Text-to-image generation.}
Text-to-image generation~\cite{ramesh2021zero, nichol2021glide, saharia2022photorealistic, rombach2022high, ramesh2022hierarchical, chen2023pixart, zhou2023shifted, esser2024scaling, dai2023emu} has rapidly advanced with diffusion and later \FM~models.
This task bridges two critical and complex domains: language and vision. 
Existing methods typically integrate text encoders, such as LLMs, into diffusion models through additional conditioning mechanisms, with cross-attention being the most prevalent~\cite{esser2024scaling, polyak2024movie}. 
However, these approaches increase model complexity and require extra parameters.
We demonstrate that \OURS{} improves over standard \FM with better scaling characteristics, and is comparable to prior work, despite a simpler architecture.

\noindent\textbf{Cross-modal / intra-modal mapping.}
Various tasks can be framed as cross-modal/intra-modal mapping problems, including image captioning~\cite{li2022blip, you2016image, gao2019masked, lee2018deterministic, guo2020non, zhou2021semi, luo2023semantic}, depth estimation~\cite{yang2024depth, ranftl2020towards, ke2024repurposing, bhat2021adabins, li2023depthformer, li2024binsformer, duan2023diffusiondepth}, and image super-resolution~\cite{saharia2022image,fischer2023boosting}.
However, due to the significant differences between modalities or distributions, previous methods have typically relied on task-specific designs.
For example, Bit Diffusion~\cite{chen2022analog} encodes text into binary bits and uses a diffusion model with self-conditioning for captioning.
Flow-based super-resolution models, such as CFM~\cite{fischer2023boosting}, still require the low-resolution image as extra conditioning, and also add Gaussian noise to the input.
In contrast, our \OURS uses the same unified framework across all these tasks without extra conditioning or noise.

\section{Preliminaries}

\begin{figure*}[ht]
    \centering
    \includegraphics[width=\linewidth]{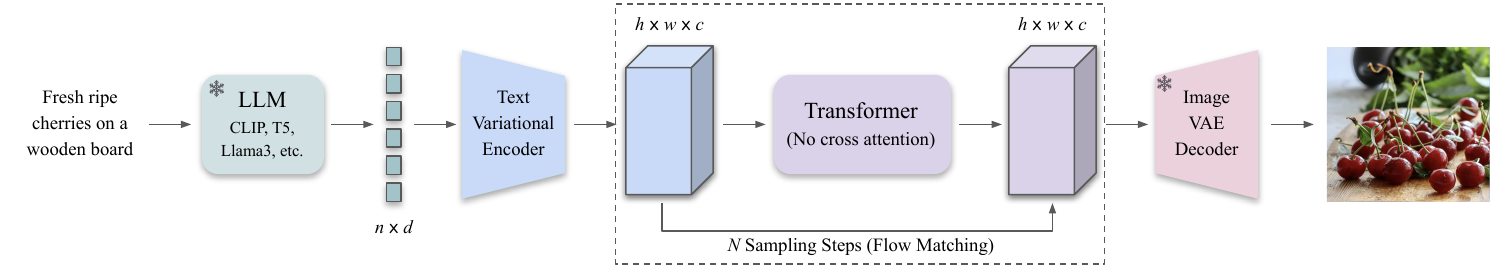}
    \vspace{-7mm}
    \caption{\textbf{\OURS Architecture.}
    \OURS enables direct evolution between two different modalities.
    Taking \ttoi generation as an example, our T2I model comprises two main components: a Text \VE and a standard \FM model.
    At inference time, we utilize the Text \VE to extract the text latent $z_0\in\mathbb{R}^{h \times w \times c}$ from text embedding $x\in\mathbb{R}^{n\times d}$ produced by any language model.
    Then we directly evolve this text latent into the image space to generate image latent $z_1\in\mathbb{R}^{h \times w \times c}$.
    }
    \vspace{-3mm}
    \label{fig:arch}
\end{figure*}

\noindent\textbf{\FMCaps.} 
We consider a generative model that defines a mapping between samples $z_0$ from a source distribution $p_0$ to samples $z_1$ of a target distribution $p_1$ via the ordinary differential equation (ODE): $dz_t=v_\theta(z_t,t)dt$.
Here, $v_\theta$ is the velocity parameterized by the weights $\theta$ of a neural network, and $t\in[0,1]$ is the time-step.
Specifically, \FMCaps~\cite{lipman2022flow, albergo2023building, liu2023flow} defines the forward process as:
\begin{align}
    z_t = t z_1 + (1-(1-\sigma_{min}) t)z_0
    \label{eqn_1}
\end{align}
and $\sigma_{min}=10^{-5}$. Ground truth velocity is computed as: 
\begin{align}
    \hat{v}_t = \frac{dz_t}{dt} = z_1 - (1-\sigma_{min})z_0
    \label{eqn_2}
\end{align}
To achieve this, a network $v_\theta(z_t,t)$ is trained to predict velocity by minimizing the mean squared error (MSE) between its output and the target $\hat{v}_t$.
This constructs a continuous path between $z_0$ and $z_1$ at any time-step $t\in[0,1]$.

As discussed earlier, \FM enables evolving a sample $z_1$ from an arbitrary source distribution $p_0$.
But prior work~\cite{esser2024scaling, polyak2024movie} typically starts from Gaussian noise $z_0\sim\mathcal{N}(0,1)$, and computing the velocity with additional condition $c$ incorporated through various methods, \eg, cross-attention~\cite{esser2024scaling, polyak2024movie}, channel-wise concatenation~\cite{girdhar2023emu}.

\noindent\textbf{\CFG.}
\CFGShort~\cite{ho2022classifier} is a broadly used technique that enhances sample quality in \emph{conditional} generative models by jointly training a single model on conditional and unconditional objectives.
This is achieved through randomly dropping the condition $c$ during training with a certain probability $p$.
Sampling is performed by extrapolating between conditional and unconditional denoising $v_\theta(z_t,c)$ and $v_\theta(z_t)$ with a scaling factor $\omega$:
\begin{align}
    \tilde{v}_\theta(z_t,c) = \omega v_\theta(z_t,c) + (1-\omega) v_\theta(z_t)
    \label{eqn_3}
\end{align}
It significantly improves the generation quality and fidelity by guiding the samples towards higher likelihood of the condition $c$, which plays a crucial role in \sota media generation models~\cite{esser2024scaling,chen2023pixart,ramesh2022hierarchical,polyak2024movie}.

\section{\OURS}
\label{sec:method}

In this section, we discuss the various components of our approach: a \VE (\VEShort) to encode the inputs in \cref{sec:ve}, using \fm to evolve from the source to the target distribution in \cref{sec:training_ours}, and finally, applying \CFGShort in this setting for improving quality and fidelity in \cref{sec:cfg_ind}.

\subsection{\VE{} for Encoding Inputs}
\label{sec:ve}
Flow matching models require the source distribution $p_0$ to have the same shape as the target distribution $p_1$. 
In particular, given an input $x$, we need to convert it to the source latent $z_0$, which has the same shape as the target latent $z_1$.
An intuitive solution is to use an encoder $\mathcal{E}$ to convert $x$ to $z_0$, \ie, $z_0=\mathcal{E}(x)$, which can preserve most of the input information as shown in \cref{sec:supp:abl}.
However, directly evolving from $\mathcal{E}(x)$ to $z_1$ is problematic. 
We find that it is essential to formulate $z_0$ as a regularized distribution for the source in order for \FM to work well. 
To address this, we propose using a \VEShort to convert $x$ to $z_0$. 
Formally, instead of directly predicting $z_0$, we predict its mean $\bar{\mu}_{z_0}$ and variance $\bar{\sigma}_{z_0}$, and then sample the latent $z_0\sim\mathcal{N}(\bar{\mu}_{z_0}, \bar{\sigma}_{z_0}^2)$.
This enables us to convert the given input $x$ into latent $z_0$ with a regularized distribution, which can then be gradually evolved into the target distribution $z_1$ with \FM.

The \VEShort can be trained with a standard Variational Autoencoding objective (VAE)~\cite{kingma2013auto} comprising of an encoding loss and the KL-divergence loss.
For the encoding loss, the \VEShort is trained to minimize a loss between the output $z_0$ and a target $\hat{z}$.
For a VAE this loss would be a reconstruction loss like MSE between the input $x$ and the decoder $\mathcal{D}${}'s output, $\texttt{MSE}(\mathcal{D}(z_0),x)$.
But since we simply need a encoder and not an autoencoder, we don't restrict ourselves to a VAE.

\subsection{Training \OURS}
\label{sec:training_ours}
For each training sample, we start with an input-target pair $(x, z_1)$.
We apply the \VEShort{} to $x$ to encode it to a latent $z_0$ with the same shape as $z_1$.
Next, we employ a transformer model $v_\theta$ trained for \FM as per Equations \ref{eqn_1} and \ref{eqn_2}.
The \VEShort can be trained prior to training $v_\theta$ or concurrently.
We show in \cref{subsec:exp_abl} that jointly training the \VE with \FM results in improved performance. 

Specifically, we jointly train the \VEShort with the \FM model using a sum of \FM MSE loss $L_{FM}$, and the losses for \VE training (encoding loss $L_{Enc}$ and KL-divergence loss $L_{KL}$):
\begin{align}
    L &= L_{FM} + L_{Enc} + \lambda L_{KL} \notag\\
    &=\texttt{MSE}(v_\theta(z_t,t),\hat{v}) + \texttt{Enc}(z_0,\hat{z}) \notag\\
    &\qquad + \lambda \texttt{KL}(\mathcal{N}(\bar{\mu}_{z_0}, \bar{\sigma}_{z_0}^2)||\mathcal{N}(0, 1))
    \label{eqn_5}
\end{align}
where $\lambda$ is the KL-divergence loss weight. 
Eq.~\ref{eqn_5} outlines the general form of the loss function across tasks, where $L_{enc}$ varies by task.
\cref{subsec:DM_T2I} discusses \ttoi generation and choices for $L_{Enc}$.
More details in \cref{sec:supp:loss_fun}.

\subsection{Classifier-Free Guidance with an Indicator}
\label{sec:cfg_ind}
\CFGShort~\cite{ho2022classifier} has become the standard low-temperature sampling method for enhancing multi-modal alignment and improving quality.
However, it can only be applied to generation methods that accept an additional conditioning input $c$, since the guidance signal relies on the difference between conditional and unconditional predictions $v_\theta(z_t,c)$ and $v_\theta(z_t)$.
Recently, Autoguidance (AG)~\cite{karras2024guiding} has been introduced as a method to improve both conditional and unconditional generation, by guiding with a smaller, less-trained `bad model'.
However, it underperforms compared to standard \CFGShort. AG also requires training a separate bad model, and its performance varies dramatically based on the choice of the bad model. 
While using an under-trained version of the same model narrows the search space, it affects performance and is impractical for large models due to the need to load two models during inference.

We instead aim to support \CFGShort for \OURS, which is as accessible and performant as \CFGShort is for standard \FM. To enable \CFGShort without the presence of an explicit conditioning input $c$, we introduce \CFGShort with indicator.
Specifically, our model is of the form $v_\theta(z_t, 1_c)$, where $1_c \in \{0, 1\}$ is an indicator to specify conditional \vs unconditional generation. 
The model evolves from $z_0$ to $z_1$ when $1_c = 1$, and from $z_0$ to $z_1^{uc}$ when $1_c = 0$, where $z_1^{uc}$ represents any sample from the target distribution $p_1$ other than $z_1$. 
During training, we employ two learnable parameters, $g^c$ and $g^{uc}$, corresponding to conditional and unconditional generation, respectively.
Depending on $1_c$, the appropriate learnable parameter is concatenated with the transformer input tokens along sequence dimension. 
We randomly sample the indicator with an unconditional rate of $10\%$, as per standard practice.
The insight behind the \CFGShort indicator is similar to that of standard \CFGShort.
In this approach, $v_\theta(z_t, 1)$ is trained to map $z_0$ to a specific region of the target manifold, while $v_\theta(z_t, 0)$ is trained to map $z_0$ to the entire target manifold to generate arbitrary unrelated images.

\subsection{Flowing from Text to Image}
\label{subsec:DM_T2I}
Now, we consider \ttoi generation as the archetypal task to leverage \OURS.  
We start with the input text embedding $x\in\mathbb{R}^{n\times d}$ with token length $n$ and dimension $d$, and use our Text \VEShort to extract the corresponding text latent $z_0\sim\mathcal{N}(\bar{\mu}_x, \bar{\sigma}_x^2)$. While our approach is agnostic to pixel \vs latent image generation, we consider image generation in the latent space for efficiency, and leverage a pre-trained VAE to obtain the image latent from the input image $\text{I}$, which serves as our target $z_1$.
Then, we employ the vanilla \FM~\cite{lipman2022flow} model to predict $v(z_t,t)$ between $z_0$ and $z_1$. The pipeline for performing \ttoi generation with \OURS is illustrated in~\cref{fig:arch}. We discuss how to train a performant Text \VE next.

\subsubsection{Text Variational Encoder}
\label{subsec:VTE}
Training the Text \VEShort is challenging, as this involves compressing the text embeddings to small latent space (\eg, $77\times768$ \CLIP tokens to $4\times32\times32$ image latents for $256$px generation, $14.4\times$ compression).
We explore various methods to train \VEShort{}s for \OURS.
The straightforward approach is to simply train a VAE with a MSE reconstruction loss.
While this approach achieves very low reconstruction errors, we find that it does not capture semantic concepts well, leading to sub-optimal image generations. 

\noindent\textbf{Contrastive loss.} %
We explore contrastive losses, which produce representations with strong semantic understanding when training on samples within the same modality~\cite{oord2018representation,chen2020simple} and on different modality pairs~\cite{radford2021learning}.
To produce the contrastive targets for the VE, we either use the input text embedding $x$ (text-text contrastive), or the paired image $I$ for the text (image-text contrastive).
Given the target, %
we employ a simple encoder to project it into a feature space with the same shape as $z_0$, resulting in a representation denoted as $\hat{z}$.
We then encourage semantic similarity between $z_0$ and $\hat{z}$ using the contrastive CLIP loss~\cite{radford2021learning}.
During training, the batch-wise contrastive loss is computed as $L_{Enc}=\texttt{CLIP}(z_0, \hat{z})$.
We ablate this choice in~\cref{subsec:exp_abl} and find that contrastive loss works significantly better than the VAE reconstruction loss, with the image-text loss working slightly better than the text-text loss.

\section{Experiments}
\label{sec:exp}

We first evaluate \OURS on \ttoi generation, demonstrate its scalability, and showcase some interesting applications with latent arithmetic in \cref{subsec:exp_ti2}. %
Then, we ablate our main design decisions through ablation studies in \cref{subsec:exp_abl}. 
Finally, we further explore \OURS{}'s performance on three distinct tasks: image captioning, monocular depth estimation, and image super-resolution in~\cref{subsec:exp_diff_mod}.

\subsection{Text-to-Image Generation}
\label{subsec:exp_ti2}

\noindent\textbf{Experimental setup.}
Scientifically comparing T2I models is challenging due to diverse training datasets, often including proprietary data, and varying training conditions.
In addition, our method represents a new paradigm for utilizing diffusion models, distinct from previous T2I approaches.
Therefore, we primarily compare our model with the widely used ``standard flow matching baseline" that starts from noise and uses text cross-attention.
For fairness, both \OURS and the baseline share the same codebase, training recipe, dataset, and budget. 
Unlike the baseline, which requires cross-attention after each self-attention layer, our model only relies on self-attention, reducing parameters per layer.
To account for this, we adjust the number of layers to match model sizes.
For both methods, we use a grid search %
to find the optimal CFG scale.
We also compare \OURS with \sota T2I models to demonstrate that our approach is competitive with those established methods. 

\noindent\textbf{Architecture.}
Our model enables the use of vanilla Transformer~\cite{vaswani2017attention} with self-attention layers and feed-forward layers. 
We use DiMR~\cite{liu2024alleviating} as the \FM backbone, a variant of Diffusion Transformer (DiT)~\cite{peebles2023scalable} which replaces the parameter-heavy MLP in adaLN-Zero with a lightweight Time-Dependent Layer Normalization.
For the Text \VEShort, we apply stacked Transformer blocks, followed by a linear layer to project the output into the target shape.

\begin{table}[]
\centering
\setlength{\lightrulewidth}{0.01em}
\setlength{\cmidrulewidth}{0.01em}
\setlength\tabcolsep{3pt}
\resizebox{0.95\columnwidth}{!}{
\begin{tabular}{l|cc|cc}
\bf Method & \bf \#Params (B) & \bf \#Steps (K) & \bf FID $\downarrow$ & \bf CLIP $\uparrow$ \\
\midrule
Standard FM (Baseline) & 1.04 & 300 & 10.79 & 0.29 \\ %
\OURS (Ours) & 0.95 & 300 & 10.13 & 0.29 \\ %
\end{tabular}
}
\vspace{-2mm}
\caption{\textbf{Comparison between our \OURS and standard \FM with cross-attention.} %
Both models are trained with the same settings. %
We find that our model slightly outperforms standard \FM baseline in terms of \textit{zero-shot} FID-30K and achieves comparable performance on the CLIP score.
}
\vspace{-2mm}
\label{tab:t2i_dm_vs_xa}
\end{table}

\noindent\textbf{Training details.}
We use a proprietary dataset with about $350$M image-text pairs to train both \OURS and our ablations.
Our text encoder is based on CLIP~\cite{radford2021learning} with a fixed sequence length of $77$ text tokens. We use a pre-trained and frozen VAE from LDM~\cite{rombach2022high} to extract image latents.
Logit-normal sampling~\cite{esser2024scaling} is used to bias the training timesteps.
All T2I models are trained using the same settings: an image resolution of $256\times 256$, a batch size of $1024$, a base learning rate of $1\times 10^{-4}$ with $5000$ warm-up steps, and an AdamW optimizer~\cite{loshchilov2017decoupled} with $\beta_1=\beta_2=0.9$ and a weight decay of $0.03$, and a KL loss weight of $\lambda=1\times 10^{-4}$.
We train our largest model ($0.95$B) on $256\times 256$ for $600$K iterations, then finetune it on $512\times 512$ for an additional $240$K iterations for higher resolution generation. 

\begin{figure}[ht]
    \centering
    \vspace{-3mm}
    \begin{subfigure}[b]{0.49\columnwidth}
        \centering
        \resizebox{\columnwidth}{!}{
            \begin{tikzpicture}
    \begin{axis}[
        xmode=log,
        xmin=50,
        xmax=1000,
        xtick={70, 150, 300, 500, 1000},
        xticklabels={70, 150, 300, 500, 1000},
        ytick={6, 10, 14, 18, 22},
        ymin=6,
        ymax=22,
        xlabel={Model parameters (millions)},
        grid=both,
        grid style={line width=.1pt, draw=gray!10},
        major grid style={line width=.2pt,draw=gray!50},
        minor tick num=2,
        axis x line*=bottom,
        axis y line*=left,
        height=2in,
        ylabel style= {align=center, yshift=-10pt},
        ylabel={FID-30K},
        yticklabel style = {font=\small},
        xticklabel style = {font=\small},
    ]

    \addplot[mark=o, very thick, snsBlueColor] plot coordinates {
        (70, 17.67)
        (150, 15.64)
        (300, 14.33)
        (500, 12.38)
        (1000, 10.79)
    };
    \addplot[mark=o, very thick, snsGreenColor] plot coordinates {
        (70, 19.48)
        (150, 17.91)
        (300, 15.07)
        (500, 12.59)
        (1000, 10.13)
    };
    \end{axis}
\end{tikzpicture}
        }
    \end{subfigure}
    \hfill
    \begin{subfigure}[b]{0.49\columnwidth}
        \centering
        \resizebox{\columnwidth}{!}{
            \begin{tikzpicture}
    \begin{axis}[
        xmin=50,
        xmax=300,
        xtick={50, 100, 150, 200, 250, 300},
        xticklabels={50, 100, 150, 200, 250, 300},
        ytick={12, 13, 14, 15, 16, 17},
        ymin=12,
        ymax=17,
        xlabel={Training steps (thousands)},
        grid=both,
        grid style={line width=.1pt, draw=gray!10},
        major grid style={line width=.2pt,draw=gray!50},
        minor tick num=2,
        axis x line*=bottom,
        axis y line*=left,
        height=2in,
        ylabel style= {align=center, yshift=-10pt},
        ylabel={FID-30K},
        yticklabel style = {font=\small},
        xticklabel style = {font=\small},
        legend style={cells={align=left}, font=\small, anchor=east},
    ]

    \addplot[mark=o, very thick, snsBlueColor] plot coordinates {
        (50, 19.65)
        (100, 15.29)
        (150, 14.59)
        (200, 13.96)
        (250, 13.62)
        (300, 13.26)
    };
    \addlegendentry{\BASELINEShort}
    \addplot[mark=o, very thick, snsGreenColor] plot coordinates {
        (50, 24.94)
        (100, 15.75)
        (150, 14.58)
        (200, 13.86)
        (250, 13.33)
        (300, 12.84)
    };
    \addlegendentry{\OURS}
    \end{axis}
    
\end{tikzpicture}
        }
    \end{subfigure}
    \vspace{-3mm}
    \caption{\textbf{Performance \vs Model Parameters and Iterations.} We compare the baseline of starting from noise with text cross-attention with \OURS, while controlling for data, model size and training steps. \emph{Left}: Larger models are able to exploit the cross-modality connection better. \emph{Right}: \OURS needs more steps to converge, but converges to better final performance. Overall, \OURS scales better than the baseline and can serve as the framework for future media generation models. }
    \vspace{-3mm}
    \label{fig:scaling}
\end{figure}

\noindent\textbf{Evaluation metrics.}
We evaluate all models on the COCO validation set~\cite{lin2014microsoft} and report FID~\cite{heusel2017gans} and CLIP score~\cite{radford2021learning, hessel2021clipscore}.
Following previous works, we report \textit{zero-shot} FID-30K, where 30K prompts are randomly sampled from the validation set, and the generated images are compared to reference images from the full validation set.
Additionally, we also evaluate our models on GenEval benchmark as it exhibits strong alignment with human judgment~\cite{ghosh2024geneval}.

\subsubsection{\OURS \textbf{\textit{vs}}. \BASELINE}
\label{sec:ours_vs_baseline}
We compare our \OURS with widely used cross-attention baseline in Tab.~\ref{tab:t2i_dm_vs_xa}.
Both models are trained and tested under the same settings. 
The results show that \OURS achieves comparable performance, with slightly better \textit{zero-shot} FID-30K compared with widely used \FM baselines with cross-attention. 

\noindent\textbf{Scaling characteristics.}
We investigate the scalability of \OURS in~\cref{fig:scaling} and compare it with standard \FM.
We train both approaches across 5 different model sizes, ranging from $70$M to $1$B parameters, with the same training settings, for $300$K iterations. 
At smaller scales, \OURS underperforms the baseline, likely due to the lack of sufficient parameters to model the complex relationships between two modalities. 
But excitingly, as the model size increases, the \textit{zero-shot} FID-30K improves more for our approach.
Next, we evaluate the effect of varying the training iterations. We notice similarly that \OURS improves more as we increase training iterations.

While \OURS initially underperforms standard \FM at small scales, increasing the model size and training iterations improves it significantly, even enabling it to surpass standard \FM.
We attribute this to the fact that \OURS generates images by directly evolving from the source distribution where different sub-regions correspond to different semantics.
In contrast, standard \FM may generate the same semantics from the entire source distribution, while exploiting the inductive biases afforded by text cross-attention.
Ultimately, this works in favor of \OURS{}, as the learnt cross-modal paths and fewer inductive biases result in improved scaling characteristics with both model size and training iterations.

\begin{table}[]
\centering
\resizebox{0.8\columnwidth}{!}{
\begin{tabular}{l|c|HHcc}
\multirow{2}{*}{\bf Method} & \multirow{2}{*}{\bf \#Params.} & \multirow{2}{*}{\bf \#Images} & \multirow{2}{*}{GPU days} & \bf FID-30K $\downarrow$ & \bf GenEval  $\uparrow$ \\
 & & & & \bf \textit{zero-shot} & \bf score\\
\midrule
DALL·E~\cite{ramesh2021zero} & 12.0B & 250M & - & 27.50 & - \\
GLIDE~\cite{nichol2021glide} & 5.0B & 250M & - & 12.24 & -\\
LDM~\cite{rombach2022high} & 1.4B & 400M & - & 12.63 & -\\
DALL·E 2~\cite{ramesh2022hierarchical} & 6.5B & 650M & 41,667 A100 & 10.39 & 0.52 \\
LDMv1.5~\cite{rombach2022high} & 0.9B & 2000M & 6,250 A100 & 9.62 & 0.43 \\
Imagen~\cite{saharia2022photorealistic} & 3.0B & 860M & 7,132 A100 & 7.27 & - \\
RAPHAEL~\cite{xue2024raphael} & 3.0B & 5000M+ & 60,000 A100 & 6.61 & - \\
PixArt-$\alpha$~\cite{chen2023pixart} & 0.6B & 25M & 753 A100 & 7.32 & 0.48\\
LDMv3 ($512^2$)~\cite{esser2024scaling} & 8.0B & & & - & 0.68 \\
\midrule
\OURS  & 0.95B & 350M & - & 9.63 & 0.55  \\ 
\OURS~(Sin-Cos) & 0.95B & 350M & - & 8.95 & 0.57  \\ 
\end{tabular}
}
\vspace{-2mm}
\caption{\textbf{Comparison to recent T2I models.} 
For GenEval, we report overall scores here and task-specific results in \cref{sec:supp:GenEval}.
\OURS achieves comparable results to \sota models by evolving text directly into images. 
\OURS~(Sin-Cos) replaces simple linear flow matching with sin-cos matching~\cite{albergo2023stochastic2}.
}
\label{tab:t2i_sota}
\end{table}

\subsubsection{\Sota Comparison}
\label{sec:sota_comparison}
Finally, we compare \OURS with \sota \ttoi models and report results in Tab.~\ref{tab:t2i_sota}. 
We additionally explore sin-cos matching~\cite{albergo2023stochastic2} and find it improves over vanilla linear flow matching.
We achieve a \textit{zero-shot} FID-30K of $8.95$ on COCO, and a GenEval score of $0.57$, demonstrating performance comparable with the \sota. 
Note that our model uses only 630 A100 GPU-days for training, whereas other methods like DALL·E 2~\cite{ramesh2022hierarchical} typically require thousands of A100 GPU days.
These results suggest that \OURS is a simple and promising direction for \sota media generation. %

\begin{figure*}[ht]
    \centering
    \includegraphics[width=\linewidth]{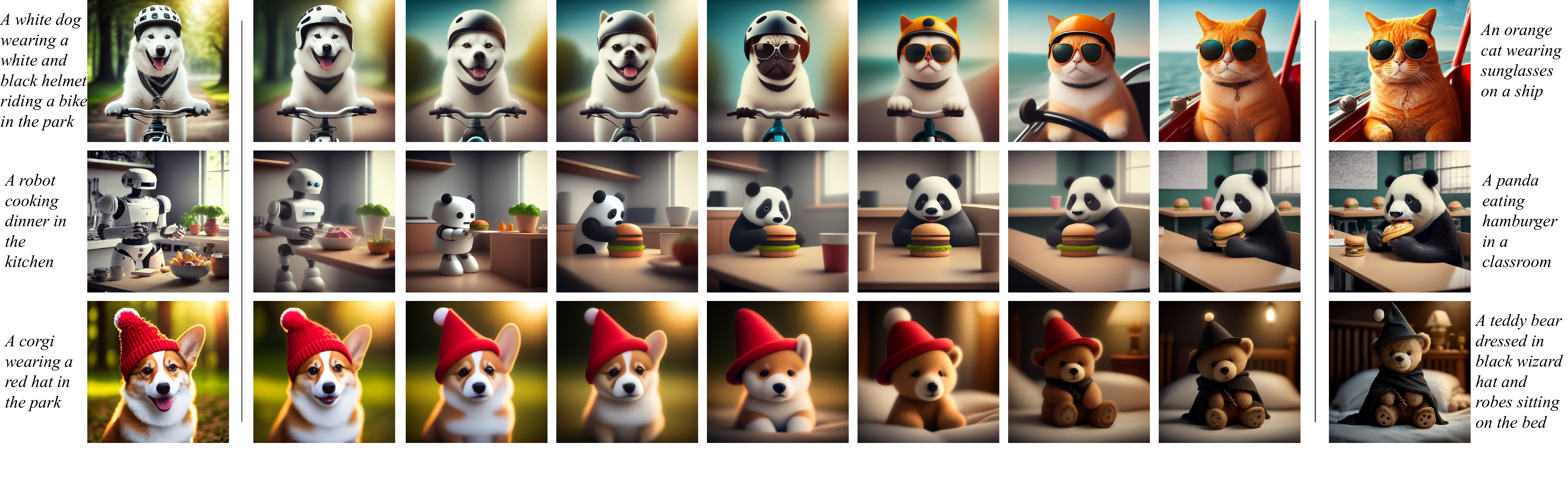}
    \vspace{-10mm}
    \caption{\textbf{\OURS provides visually smooth interpolations in the latent space.}
    We show images generated by linear interpolation between the first (left) and second (right) text latents.
    \OURS enables visually smooth transformations of object direction, composite colors, shapes, background scenes, and even object categories.
    Please zoom in for better visualization.
    For brevity, we display only 7 interpolating images here; additional interpolating images can be found in \cref{sec:supp:addl_qual} (\cref{fig:supp_interp_2} and \cref{fig:supp_interp_3}).
    }
    \vspace{-3mm}
    \label{fig:interpolation}
\end{figure*}

\begin{figure}[ht]
    \centering
    \includegraphics[width=\columnwidth]{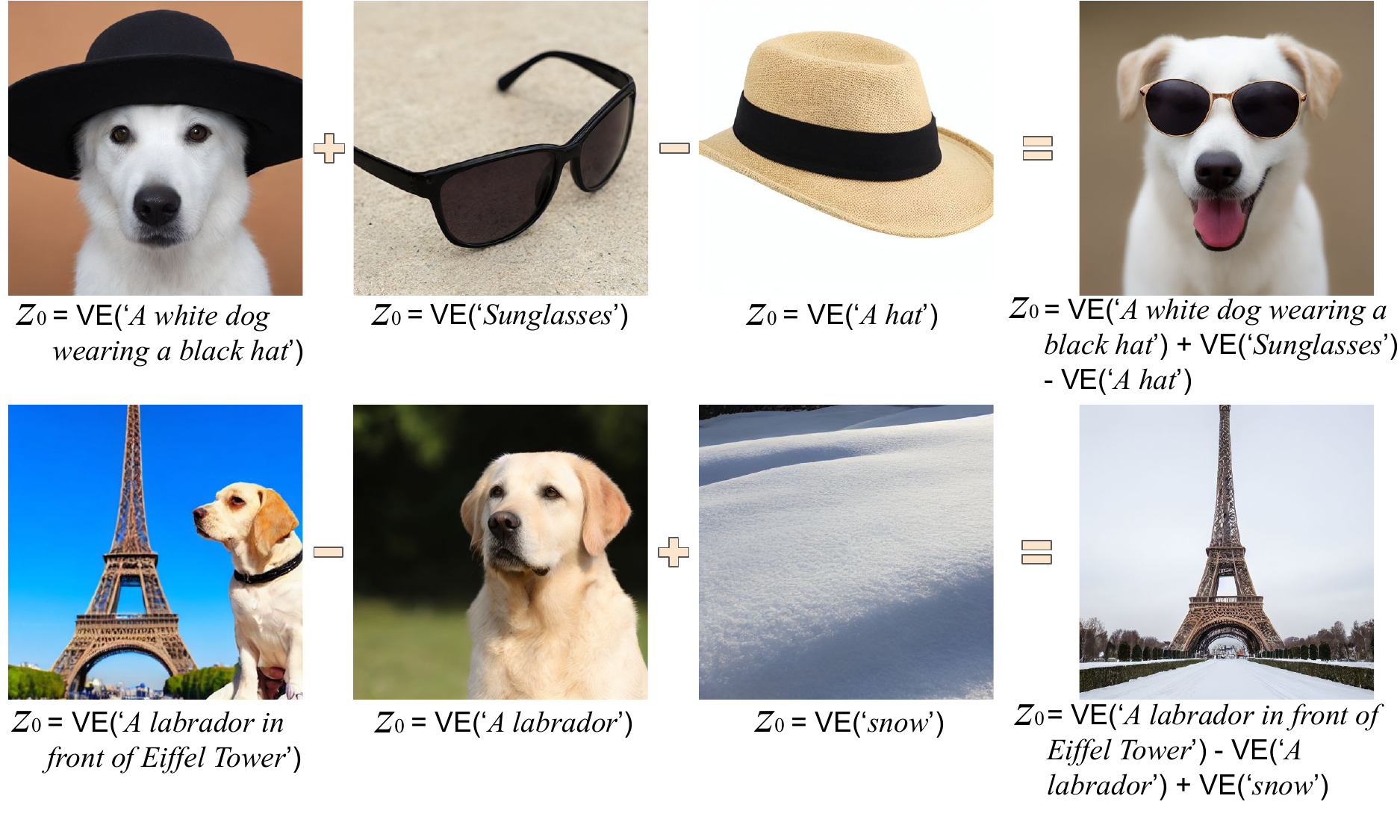}
    \vspace{-10mm}
    \caption{\textbf{\OURS allows arithmetic in text latent space.}
    Using the Text \VE (VE), we first map the input text into the latent space $z_0$. 
    Arithmetic operations are then performed in this latent space, and the resulting latent representation is used to generate the corresponding image.
    The latent code $z_0$ used to generate each image is provided at the bottom.
    }
    \vspace{-2mm}
    \label{fig:arithmetic}
\end{figure}

\subsubsection{Arithmetic Operations in Latent Space}
\label{sec:exp_int}
Unlike previous diffusion or \FM models, \OURS offers a unique property: arithmetic operations in the input latent space translate to similar operations in the output space.
This is made possible since \OURS transforms the source space (\ie, the text latent space for T2I) into a regularized continuous space, where a uniform representation shape is shared across all texts.
We showcase two examples of this, latent interpolation, and latent arithmetic. 
For latent interpolation, we use the Text \VE to generate text latents from two different text inputs, and then interpolate between them to produce images.
As shown in Fig.~\ref{fig:interpolation}, \OURS enables visually smooth linear interpolations, even between disparate prompts.
Next, we showcase arithmetic operations in Fig.~\ref{fig:arithmetic}, in which we apply addition and subtraction in the text latent space, and find that the resulting images exhibit corresponding semantic modifications to the original image.
This shows that \OURS formulates meaningful and well-structured semantic paths between the source and target distributions, providing additional capabilities and more control over standard \FM approaches.
See \cref{sec:supp:Arithmetic} for further details.

\subsection{Ablation Study}
\label{subsec:exp_abl}

We conduct various ablation experiments to verify the effectiveness of the proposed designs in Tab.~\ref{tab:abl_main}.

\begin{table}[ht]
  \centering
  \subfloat[\bf \scriptsize \VE\textsuperscript{*}]{
    \resizebox{0.55\linewidth}{!}{
      \begin{tabular}{l|cc}
        \bf Text encoder & \bf FID $\downarrow$ & \bf CLIP $\uparrow$ \\
        \midrule
        Encoder & 66.65 & 0.20 \\
        Encoder + noise & 59.91 & 0.21 \\
        \underline{\VE} & \underline{40.78} & \underline{0.23} \\
      \end{tabular}
    }
  }
  \subfloat[\bf \scriptsize Text \VEShort loss\textsuperscript{*}]{
    \hspace{-3mm}
    \resizebox{0.47\linewidth}{!}{
      \begin{tabular}{l|cc}
        \bf Loss & \bf FID $\downarrow$ & \bf CLIP $\uparrow$ \\
        \midrule
        T-T Recon. &  40.78 & 0.23 \\
        T-T Contrast. & 34.67 & 0.24  \\
        \underline{I-T Contrast.} & \underline{33.41} & \underline{0.24}  \\
      \end{tabular}
    }
  } \\
  \vspace{1mm}
  \subfloat[\bf \scriptsize CFG with indicator]{
    \resizebox{0.48\linewidth}{!}{
      \begin{tabular}{l|cc}
        \bf Method  & \bf FID $\downarrow$ & \bf CLIP $\uparrow$ \\
        \midrule
        No guidance & 33.41 & 0.24 \\
        AG  & 26.36 & 0.25 \\
        \underline{CFG indicator} & \underline{24.33} & \underline{0.26} \\
      \end{tabular}
    }
  } 
  \subfloat[\bf \scriptsize Language Model]{
    \resizebox{0.5\linewidth}{!}{
        \begin{tabular}{lH|cc}
        \bf Model & \bf \#Params & \bf FID $\downarrow$ & \bf CLIP $\uparrow$ \\
        \midrule
        \underline{CLIP (0.4B)} & \underline{0.4B} & \underline{24.33} & \underline{0.26} \\
        T5-XXL (11B) & 11B & 22.28 & 0.27 \\
        \llamathree (7B) & 7B & 21.20 & 0.27 \\
        \end{tabular}
    }
  } \\
  \vspace{1mm}
  \subfloat[\bf \scriptsize Training strategy]{
    \resizebox{0.67\linewidth}{!}{
      \begin{tabular}{l|cc}
        \bf Train strategy & \bf FID $\downarrow$ & \bf CLIP $\uparrow$ \\
        \midrule
        2-stage separate training & 32.55 & 0.24 \\
        \underline{Joint training} & \underline{24.33} & \underline{0.26} \\
        2-stage w/ joint finetuning & 23.79 & 0.26 \\
      \end{tabular}
    }
  }
\vspace{-3mm}
\caption{\textbf{Ablation study} on Text \VE, training objective, CFG, language models, and training strategy. 
We conduct ablation study on our smallest model (70M), reporting \textit{zero-shot} FID-10K and CLIP scores.
Final settings used for \OURS are \underline{underlined}. AG: Autoguidance. \textsuperscript{*}: results without applying CFG.}
\vspace{-1mm}
\label{tab:abl_main}
\end{table}

\noindent\textbf{Variational Encoder \vs standard encoder.}
Compared to a standard encoder or even adding Guassian noise like CFM~\cite{fischer2023boosting}, a \VE significantly improves the generation quality, with significant gains in the FID.
This shows that forming a regularized distribution for the source domain is a crucial step for \XFM.

\noindent\textbf{Joint training \vs two-stage training.}
We consider three training strategies: (1) jointly training the \VEShort and \FM from scratch, (2) training the \VEShort first and then training \FM with a fixed \VEShort, and (3) training the \VEShort first and then training the \FM while jointly fine-tuning \VEShort.
We observe that it is important to update the \VEShort when training the \FM, either through joint training from scratch, or finetuning the \VEShort jointly with \FM. Initializing with a pre-trained \VEShort and then jointly training improves convergence speed by about $35\%$, but we opt to jointly train both models from scratch on account of the simplicity, and for fair comparisons with baselines.

\noindent\textbf{CFG indicator.}
We evaluate the performance of our model when leveraging our proposed \CFGShort indicator techinuqe. We also evaluate Autoguidance (AG)~\cite{karras2024guiding}, which utilizes two models for inference -- we use an under-trained version of the same model as the bad model, while using a grid-search to find the best under-trained checkpoint. 
While AG improves FID and also image-text CLIP alignment slightly, our CFG indicator works better than AG in terms of both FID and CLIP alignment while only using a single model trained with standard \CFGShort settings. 
Qualitatively, our approach produces much higher fidelity images compared to both alternatives, as shared in \cref{sec:supp:abl}.

\noindent\textbf{Text \VEShort loss.}
We explore reconstruction and contrastive objectives for the encoder loss $L_{Enc}$ when training the text \VEShort.
We find that contrastive loss, which enhances semantic understanding, significantly outperforms reconstruction loss on input text embeddings. 
Moreover, image-text contrastive loss slightly surpasses text-text contrastive loss.

\noindent\textbf{Effect of different language models.}
We evaluate \OURS with various language models trained with different objectives. 
Specifically, we evaluate CLIP~\cite{radford2021learning} (contrastive image-text), T5-XXL's encoder~\cite{raffel2020exploring} (encoder-decoder), \llamathree-7B~\cite{dubey2024llama} (decoder-only).
We use $77$ tokens for all language models, resulting in text embeddings of size $77\times768$, $77\times4096$, $77\times4096$, respectively.
We train a separate Text \VEShort for each language model, projecting the text embeddings into the target image latent shape ($4\times32\times32$).
\OURS works well with all language models regardless of their training objectives and embedding sizes. 
As expected, our performance improves with better text representations. %
Due to compute restrictions however, we use the light-weight CLIP model for our main experiments.

\subsection{\OURS for Various Tasks}
\label{subsec:exp_diff_mod}
We further evaluate \OURS on three distinct tasks that involve cross-modal / intra-modal evolution.
We present the main results and key findings here, while additional details and qualitative results can be found in the Appendix.

\noindent\textbf{Image to text (captioning).}
We first consider the task of image captioning.
To achieve this, we train a new Text \VE on the captioning dataset to extract text latents from text tokens, and a separate text decoder with a reconstruction loss to convert text latents back into tokens.
\OURS is then trained to map from the image latent space to the text latent space. 
Following previous work, we use the Karpathy split~\cite{karpathy2015deep} of COCO dataset~\cite{lin2014microsoft} for training and testing. %
In addition, we can also leverage the bi-directional flow property, and simply fine-tune our text-to-image \OURS~model on COCO and use its inversion for captioning.
We report results in Tab.~\ref{tab:cap}.
\OURS enables direct evolution from image space to text space for image captioning, achieving \sota performance.

\begin{table}[]
\centering
\resizebox{0.85\columnwidth}{!}{
\begin{tabular}{l|ccccc}
\bf Method & \bf B@4 $\uparrow$ & \bf M $\uparrow$ & \bf R $\uparrow$ & \bf C $\uparrow$ & \bf S $\uparrow$ \\
\midrule
MNIC~\cite{gao2019masked} & 30.9 & 27.5 & 55.6 & 108.1 & 21.0 \\
MIR~\cite{lee2018deterministic} & 32.5 & 27.2 & - & 109.5 & 20.6 \\
NAIC-CMAL~\cite{guo2020non} & 35.3 & 27.3 & 56.9 & 115.5 & 20.8 \\
SATIC~\cite{zhou2021semi} & 32.9 & 27.0 & - & 111.0 & 20.5 \\
SCD-Net~\cite{luo2023semantic} & 37.3 & 28.1 & 58.0 & 118.0 & 21.6 \\
\midrule
\OURS-T2I (Ours) & 33.1 & 27.0 & 56.4 & 111.2 & 20.3 \\
\OURS (Ours) & 36.4 & 27.8 & 57.1 & 116.2 & 20.4 \\
\end{tabular}
}
\vspace{-2mm}
\caption{\textbf{Image captioning on COCO Karpathy split.}
\OURS directly evolves from image to text, achieving comparable performance to state-of-the-art models on image captioning.
For a fair comparison, we consider non-autoregressive methods that are trained without CIDEr optimization.
\OURS-T2I achieves captioning by simply inverting our text-to-image \OURS model.
}
\label{tab:cap}
\end{table}

\noindent\textbf{Image to depth (depth estimation).}
For monocular depth estimation, we train \OURS in pixel space. 
Specifically, we use a recontruction loss to train the Image \VE to map the original image into the shape of a depth map, followed by the \FM model which generates the final depth maps. We train and evaluate our model %
on KITTI~\cite{geiger2013vision} (Eigen split~\cite{eigen2014depth}) and NYUv2~\cite{silberman2012indoor} (official split) for outdoor and indoor scenarios, respectively.
As shown in Tab.~\ref{tab:depth}, our model achieves comparable performance to state-of-the-art methods on both datasets.
Notably, DiffusionDepth~\cite{duan2023diffusiondepth} utilizes Swin Transformer~\cite{liu2021swin} and specific designs such as Multi-Scale Aggregation and Monocular Conditioned Denoising Block.
In contrast, our model achieves similar performance without any additional enhancements, demonstrating the efficiency and effectiveness of \OURS in mapping from images to depth. 
\begin{table}[]
\centering
\resizebox{0.93\columnwidth}{!}{
\begin{tabular}{l|cc|cc}
\multirow{2}{*}{\bf Method} & \multicolumn{2}{c|}{\bf KITTI} & \multicolumn{2}{c}{\bf NYUv2} \\ \cmidrule(lr){2-3}  \cmidrule(lr){4-5}
                        & \bf AbsRel ($\downarrow$)& $\boldsymbol{\delta_1}$ ($\uparrow$) & \bf AbsRel ($\downarrow$)& $\boldsymbol{\delta_1}$ ($\uparrow$)  \\ 
\midrule
TransDepth~\cite{yang2021transformer} & 0.064 & 0.956 & 0.106 & 0.900 \\
AdaBins~\cite{bhat2021adabins} & 0.058 & 0.964 & 0.103 & 0.903 \\
DepthFormer~\cite{li2023depthformer} & 0.052 & 0.975 & 0.096 & 0.921 \\
BinsFormer~\cite{li2024binsformer} & 0.052 & 0.974 & 0.094 & 0.925 \\
DiffusionDepth~\cite{duan2023diffusiondepth}  & 0.050 & 0.977 & 0.085 & 0.939  \\
\midrule
\OURS (Ours)  & 0.053 & 0.973 & 0.094 & 0.928 \\ 
\end{tabular}}
\vspace{-2mm}
\caption{\textbf{Monocular depth estimation on KITTI and NYUv2.} 
\OURS enables direct mapping from image to depth, achieving comparable performance to \sota models.
}
\vspace{-1mm}
\label{tab:depth}
\end{table}

\noindent\textbf{Low-resolution to high-resolution (super-resolution).}
We compare \OURS~with the standard flow-matching super-resolution method, which upsamples the low-resolution image, concatenates it with input noise as conditioning, and then processes it through the neural network.
In contrast, we directly evolve the upsampled low-resolution image into a high-resolution image, without additional concatenation conditioning. We also compare against SR3~\cite{saharia2022image} which uses diffusion models for super-resolution.
Following previous work~\cite{saharia2022image, lipman2022flow}, we train and evaluate our model %
on ImageNet~\cite{deng2009imagenet} for $64\times 64 \rightarrow 256\times 256$ super-resolution, and provide results in Tab.~\ref{tab:sr}.
Our method achieves better results compared to the standard \FM and SR3, indicating that \OURS can also effectively evolve between similar distributions while achieving superior performance. 

\begin{table}[]
\centering
\resizebox{0.85\columnwidth}{!}{
\begin{tabular}{l|cccc}
\bf Method & \bf FID $\downarrow$ & \bf IS $\uparrow$ & \bf PSNR $\uparrow$ & \bf SSIM $\uparrow$ \\
\midrule
Reference & 1.9 & 240.8 & - & - \\
\midrule
Regression & 15.2 & 121.1 & 27.9 & 0.801 \\
SR3~\cite{saharia2022image} & 5.2 & 180.1 & 26.4 & 0.762 \\
Flow Matching~\cite{lipman2022flow} & 3.4 & 200.8 & 24.7 & 0.747 \\
\midrule
\OURS (Ours) & 3.0 &  207.2 & 25.6 & 0.764\\
\end{tabular}
}
\vspace{-2mm}
\caption{\textbf{Image super-resolution on the ImageNet validation set.}
Our direct mapping method achieves better performance.
}
\label{tab:sr}
\end{table}

\section{Conclusion}

In this paper, we proposed \OURS, a simple and general framework for cross-modal flow matching that works well across a variety of tasks without requiring task specific architectural modifications. It outperforms conventional flow matching, while also enabling new capabilities such as latent arithmetic. We showcase that \OURS is a promising approach for the future thanks to its better scalablity. We hope our approach helps pave the way towards further research and applications of \XFM.

\noindent\textbf{Acknowledgements.}
We sincerely appreciate Ricky Chen and Saketh Rambhatla for their valuable discussions.

{
    \small
    \bibliographystyle{ieeenat_fullname}
    \bibliography{refs}
}

\clearpage
\appendix

\section*{Appendix}
\label{sec:appendix}

In the appendix, we provide additional information as listed below:

\begin{itemize}[label={\textbf{--}}]
\item Sec.~\ref{sec:supp:method}. Method details
    \begin{itemize}[label={\textbf{--}}]
    \item Sec.~\ref{sec:supp:loss_fun}. Loss function for \ttoi generation
    \item Sec.~\ref{sec:supp:var_tasks}. Experimental details for various tasks
    \end{itemize}
\item Sec.~\ref{sec:supp:addl_exps}. Additional experimental results
    \begin{itemize}[label={\textbf{--}}]
    \item Sec.~\ref{sec:supp:GenEval}. GenEval performance for \ttoi
    \item Sec.~\ref{sec:supp:Arithmetic}. Analysis of Arithmetic Operations
    \item Sec.~\ref{sec:supp:Zero-shot}. \emph{Zero-shot} depth estimation
    \item Sec.~\ref{sec:supp:sr}. Image super-resolution
    \item Sec.~\ref{sec:supp:abl}. Ablations on Text VE and CFG indicator
    \end{itemize}
\item Sec.~\ref{sec:supp:addl_qual}. Additional qualitative examples
    \begin{itemize}[label={\textbf{--}}]
    \item Fig.~\ref{fig:supp_t2i}. Text-to-image generation
    \item Fig.~\ref{fig:supp_interp_2}, ~\ref{fig:supp_interp_3}. Interpolation in latent space
    \item Fig.~\ref{fig:supp_arithmetic}. Arithmetic in latent space
    \end{itemize}
\end{itemize}

\section{Method Details}
\label{sec:supp:method}
\subsection{Loss Function for T2I Generation}
\label{sec:supp:loss_fun}
We jointly train the Text \VE with the \FM model using the following training objective:
\begin{align}
    L &= L_{FM} + L_{Enc} + \lambda L_{KL} \notag\\
    &=\texttt{MSE}(v_\theta(z_t,t),\hat{v}) + \texttt{CLIP}(z_0, \hat{z}) \notag\\
    &\qquad + \lambda \texttt{KL}(\mathcal{N}(\bar{\mu}_{z_0}, \bar{\sigma}_{z_0}^2)||\mathcal{N}(0, 1))
\end{align}
where $\lambda$ is the weight of KL-divergence loss. 
For the \FM loss $L_{FM}$, we follow previous work~\cite{lipman2022flow} and compute the MSE loss between the predicted velocity $v_\theta(z_t,t)$ at time-step $t$ and the ground-truth velocity $\hat{v}$.
To train the Text \VE, we adopt a CLIP contrastive loss.
Specifically, given a batch of $N$ text and image pairs, we use our Text \VE to obtain text latents $z_0$, and an image encoder to extract image features $\hat{z}$.
Then, we compute the cosine similarity between all pairs of $z_0$ and $\hat{z}$ in the batch, resulting in a similarity matrix $S$, where each element $s_{ij}$ represents the cosine similarity between the $i^{th}$ $z_0$ and $j^{th}$ $\hat{z}$.
The similarity scores are then scaled by a temperature parameter $\tau$ (a learnable parameter), denoted as $\text{logit}s_{ij} = s_{ij} / \tau$.
After that, a symmetric cross-entropy loss over the similarity scores is computed:
\begin{align}
    L_{\text{I2T}} = -\frac{1}{N}\sum_{i=1}^N\text{log}\frac{\text{exp}(\text{logit}s_{ii})}{\sum_{j=1}^N\text{exp}(\text{logit}s_{ij})} \\
    L_{\text{T2I}} = -\frac{1}{N}\sum_{i=1}^N\text{log}\frac{\text{exp}(\text{logit}s_{ii})}{\sum_{j=1}^N\text{exp}(\text{logit}s_{ji})}
\end{align}
Finally, we compute the average of these two components to obtain the CLIP loss, which is then used to update our Text \VE:
\begin{align}
    L_{Enc} = \texttt{CLIP}(z_0, \hat{z}) = \frac{1}{2}(L_{\text{I2T}} + L_{\text{T2I}})
\end{align}
For the KL loss $L_{KL}$, we adopt the original KL divergence loss~\cite{kullback1951information} with $\lambda=1\times 10^{-4}$.

\subsection{Experimental Details for Various Tasks}
\label{sec:supp:var_tasks}
\noindent\textbf{Image captioning.}
We conduct our experiments on the popular Karpathy split~\cite{karpathy2015deep} of COCO dataset~\cite{lin2014microsoft}, which contains $113,287$ images for training, $5,000$ images for validation, and $5,000$ image for testing.
We train our model with $351$M parameters on the training split for 100 epochs, using a batch size of $256$ and a base learning rate of $2\times10^{-4}$ with $5$ warm-up epochs.
Following the standard evaluation setup, we compare the performance over five metrics: BLEU@4~\cite{papineni2002bleu} (B@4), METEOR~\cite{banerjee2005meteor} (M), ROUGE~\cite{lin2004rouge} (R), CIDEr~\cite{vedantam2015cider} (C), and SPICE~\cite{anderson2016spice} (S).

\noindent\textbf{Monocular depth estimation.}
We consider KITTI~\cite{geiger2013vision} and NYUv2~\cite{silberman2012indoor} for outdoor and indoor depth estimation.
For KITTI, we use the Eigen split~\cite{eigen2014depth}, consisting of $23,488$ training images and $697$ testing images.
For NYUv2, we adopt the official split, which contains $24,231$ training images and $654$ testing images.
We train our model with $527$M parameters on the corresponding training splits for $50$ epochs.
We use a batch size of $64$, and decay the learning rate from $1\times10^{-4}$ to $1\times10^{-8}$ with cosine annealing.

\noindent\textbf{Image super-resolution.}
We consider natural image super-resolution, training our model on ImageNet 1K~\cite{deng2009imagenet} for the task of $64\times 64 \rightarrow 256\times 256$ super-resolution. We use the dev split for evaluation.
During training, we preprocess the images by removing those where the shorter side is less than $256$ pixels. 
The remaining images are then centrally cropped and resized to $256\times 256$.
The low-resolution images are then generated by downsampling the $256\times 256$ images using bicubic interpolation with anti-aliasing enabled.
For a fair comparison with SR3~\cite{saharia2022image}, we train our \OURS with $505$M parameters (compared to $625$M parameters in SR3).
Our model is trained for $1$M training steps with a batch size of $512$ and a learning rate of $1\times10^{-4}$, including $5,000$ warm-up steps.

\section{Additional Experimental Results}
\label{sec:supp:addl_exps}
\subsection{GenEval Performance}
\label{sec:supp:GenEval}
\begin{table}[]
\setlength{\lightrulewidth}{0.01em}
\setlength{\cmidrulewidth}{0.01em}
\setlength\tabcolsep{3pt}
\centering
\resizebox{\columnwidth}{!}{
\begin{tabular}{l|ccccccc}
\multirow{2}{*}{\bf Method} & \multirow{2}{*}{\bf Overall} & \bf Single & \bf Two & \multirow{2}{*}{\bf Counting} & \multirow{2}{*}{\bf Colors} & \multirow{2}{*}{\bf Position}  & \bf Attribute \\
 & & \bf Object & \bf Object & & & & \bf binding \\
\midrule
DALL·E 2~\cite{ramesh2022hierarchical} & 0.52 & 0.94 & 0.66 & 0.49 & 0.77 & 0.10 & 0.19\\
LDMv1.5~\cite{rombach2022high} & 0.43 & 0.97 & 0.38 & 0.35 & 0.76 & 0.04 & 0.06 \\
LDMv2.1~\cite{rombach2022high} & 0.50 & 0.98 & 0.51 & 0.44 & 0.85 & 0.07 & 0.17 \\
LDM-XL~\cite{podell2023sdxl} & 0.55 & 0.98 & 0.74 & 0.39 & 0.85 & 0.15 & 0.23 \\
PixArt-$\alpha$~\cite{chen2023pixart} & 0.48 & 0.98 & 0.50 & 0.44 & 0.80 & 0.08 & 0.07\\
LDMv3 ($512^2$)~\cite{esser2024scaling} & 0.68 & 0.98 &  0.84 &  0.66 &  0.74 &  0.40 & 0.43 \\
\midrule
\OURS  & 0.55 & 0.98 & 0.72 & 0.39 & 0.82 & 0.18 & 0.21 \\ 
\end{tabular}
}
\caption{\textbf{GenEval comparisons.} 
Our model achieves comparable performance to \sota models such as LDM-XL and DALL·E 2, suggesting that \OURS is a simple and promising direction for \sota media generation.
}
\label{tab:t2i_geneval}
\end{table}

To compare with recent \ttoi models on GenEval, we report the overall score and task-specific scores in Tab.~\ref{tab:t2i_geneval}.
Our model achieves comparable performance to \sota models such as LDMv2.1~\cite{rombach2022high}, LDM-XL~\cite{podell2023sdxl}, and DALL·E 2~\cite{ramesh2022hierarchical}.
This demonstrates that directly evolving from text space to image space with our approach is a simple and effective solution for \ttoi generation, indicating a novel and promising direction for \sota media generation.

\subsection{Analysis of Arithmetic Operations}
\label{sec:supp:Arithmetic}
\begin{figure}[]
    \centering
    \includegraphics[width=\columnwidth]{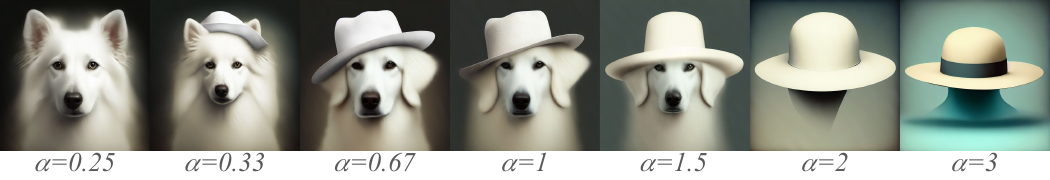}
    \vspace{-5mm}
    \caption{\textbf{Arithmetic operation with different scaling terms.} We show images generated by : $\text{VE(`a white dog')} + \alpha\text{VE(`a hat')}$ 
    }
    \label{rebuttal:fig:arith_scale}
\end{figure}

\begin{table}[]
\centering
\resizebox{0.7\columnwidth}{!}{
\begin{tabular}{l|c}
\bf Arithmetic Operation & \bf Success Rate (\%) \\
\midrule
Addition & 95.3 \\
Subtraction & 92.7 \\
Combination & 87.5 \\
\midrule
Overall & 91.4 \\
\end{tabular}
}
\caption{\textbf{Success rate of arithmetic operation}
We select 1,000 prompts from COCO-val to evaluate the success rate of arithmetic operations. 
The detection model is used to determine whether the target objects have been successfully added or removed.
``Combination'' refers to multiple operations involving a combination of both ``addition'' and ``subtraction''.
}
\label{tab:arithmetic_rate}
\end{table}

Our model encodes text into a \emph{continuous} latent space with semantic structure.
Prior work, such as word2vec~\cite{mikolov2013efficient}, has shown that latent arithmetic can \emph{emerge} without explicit training.
Arithmetic on these latents effectively retains added and removes subtracted textual information, which our Flow Matching then correspondingly maps to images. 
We analyze latent arithmetic operations in more detail here.
First, we consider addition ($+$) and test different scaling factors (Fig.~\ref{rebuttal:fig:arith_scale}), showing how they control the amount of information added or removed in the generated image.

In addition, we show qualitatively that the arithmetic works well across diverse concepts and multiple operations in Tab.~\ref{tab:arithmetic_rate}.
Specifically, we select 1,000 prompts from COCO-val to test arithmetic operations. 
A detection model confirms that \emph{91.4\% of the objects are accurately added or removed} from the generations, providing quantitative evidence of effective feature disentanglement.

\subsection{Zero-shot Depth Estimation}
\label{sec:supp:Zero-shot}
\begin{table*}[]
\centering
\resizebox{0.9\linewidth}{!}{
\begin{tabular}{l|c|cccccccccc}
\multirow{2}{*}{\bf Method} & \multirow{2}{*}{\bf \# Training samples} & \multicolumn{2}{c}{\bf KITTI} & \multicolumn{2}{c}{\bf NYUv2} &  \multicolumn{2}{c}{\bf ETH3D} & \multicolumn{2}{c}{\bf ScanNet} & \multicolumn{2}{c}{\bf DIODE} \\ 
\cmidrule(lr){3-4}  \cmidrule(lr){5-6} \cmidrule(lr){7-8} \cmidrule(lr){9-10} \cmidrule(lr){11-12}
& & \bf AbsRel $\downarrow$& $\boldsymbol{\delta_1}$ $\uparrow$ & \bf AbsRel $\downarrow$& $\boldsymbol{\delta_1}$ $\uparrow$ & \bf AbsRel $\downarrow$& $\boldsymbol{\delta_1}$ $\uparrow$ & \bf AbsRel $\downarrow$ & $\boldsymbol{\delta_1}$ $\uparrow$ & \bf AbsRel $\downarrow$ & $\boldsymbol{\delta_1}$ $\uparrow$ \\ 
\midrule
DiverseDepth~\cite{yin2020diversedepth} & 320K & 0.117 & 0.875 & 0.190 & 0.704 & 0.228 & 0.694 & 0.109 & 0.882 & 0.376 & 0.631 \\
MiDaS~\cite{ranftl2020towards} & 2M & 0.111 & 0.885 & 0.236 & 0.630 & 0.184 & 0.752 & 0.121 & 0.846 & 0.332 & 0.715 \\
LeReS~\cite{yin2021learning} & 300K + 54K & 0.090 & 0.916 & 0.149 & 0.784 & 0.171 & 0.777 & 0.091 & 0.917 & 0.271 & 0.766\\
Omnidata~\cite{eftekhar2021omnidata} & 11.9M + 310K & 0.074 & 0.945 & 0.149 & 0.835 & 0.166 & 0.778 & \textit{0.075} & 0.936 & 0.339 & 0.742 \\
HDN~\cite{zhang2022hierarchical} & 300K & \textit{0.069} & \textit{0.948} & 0.115 & 0.867 & 0.121 & 0.833 & 0.080 & \textit{0.939} & \underline{0.246} & \textbf{0.780}  \\
DPT~\cite{ranftl2021vision} & 1.2M + 188K & 0.098 & 0.903 & \textbf{0.100} & \textit{0.901} & \underline{0.078} & \underline{0.946} & 0.082 & 0.934 & \textbf{0.182} & 0.758 \\
Marigold~\cite{ke2024repurposing} & 74K & \textbf{0.060} & \textbf{0.959} & \textit{0.105} & \underline{0.904} & \textbf{0.071} & \textbf{0.951} & \underline{0.069} & \textbf{0.945} & 0.310 & \underline{0.772} \\
\midrule
\OURS (Ours) & 74K & \underline{0.062} & \underline{0.956} & \underline{0.103} & \textbf{0.908} & \textit{0.085} & \textit{0.944} & \textbf{0.068} & \underline{0.942} & \textit{0.270} & \textit{0.768} \\ 
\end{tabular}}
\caption{\textbf{\textit{Zero-shot} depth estimation.} 
Baseline results are reported by Marigold~\cite{ke2024repurposing}.
We follow Marigold and train our \OURS on the same datasets, \ie, Hypersim~\cite{roberts2021hypersim} and Virtual KITTI~\cite{cabon2020virtual}.
We highlight the \textbf{best}, \underline{second best}, and \textit{third best} entries.
With just a unified framework, \OURS achieves comparable or even superior performance on complex \emph{zero-shot} depth estimation, demonstrating the general-purpose nature of \OURS on various cross-modal tasks.
}
\label{tab:zero_shot_depth}
\end{table*}

\begin{figure}[]
    \centering
    \includegraphics[width=\columnwidth]{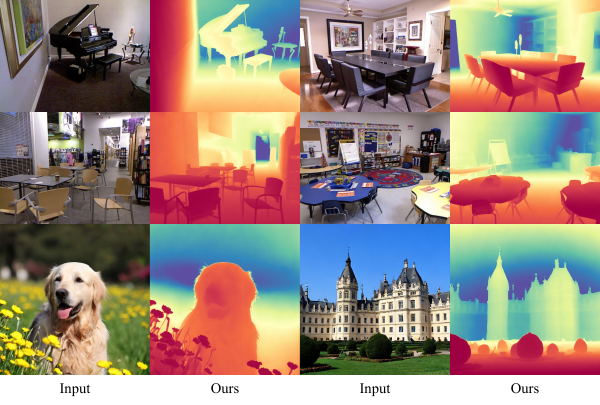}
    \vspace{-5mm}
    \caption{\textbf{Qualitative examples for \emph{zero-shot} depth estimation.}
    The input images in the first two rows are from the NYUv2 dataset, while the input images in the last row were generated by our T2I model.
    Our model provides robust \emph{zero-shot} depth estimation across domains, whether indoor or outdoor, synthetic or real.
    }
    \label{fig:depth_est_qualitative}
\end{figure}

We also evaluate \OURS on \emph{zero-shot} depth estimation.
Following Marigold~\cite{ke2024repurposing}, we train our model on Hypersim~\cite{roberts2021hypersim} and Virtual KITTI~\cite{cabon2020virtual}, and evaluate our model on 5 real datasets that are not seen during training: KITTI~\cite{geiger2013vision}, NYUv2~\cite{silberman2012indoor}, ETH3D~\cite{schops2017multi}, ScanNet~\cite{dai2017scannet}, and DIODE~\cite{vasiljevic2019diode}.
We follow Marigold~\cite{ke2024repurposing} to prepare the training and testing data.
Our model with $527$M parameters is trained for 150K training steps, with a batch size of 512 and a learning rate of $1\times10^{-4}$ with $5,000$ warm-up steps.
The results are reported in Tab.~\ref{tab:zero_shot_depth}.
Qualitative examples are provided in Fig.~\ref{fig:depth_est_qualitative}.
Without specific design, \OURS achieves comparable or even superior performance compared to \sota methods, demonstrating the general-purpose nature of our approach on various cross-modal tasks.

\subsection{Image Super-resolution}
\begin{figure}[]
    \centering
    \includegraphics[width=\columnwidth]{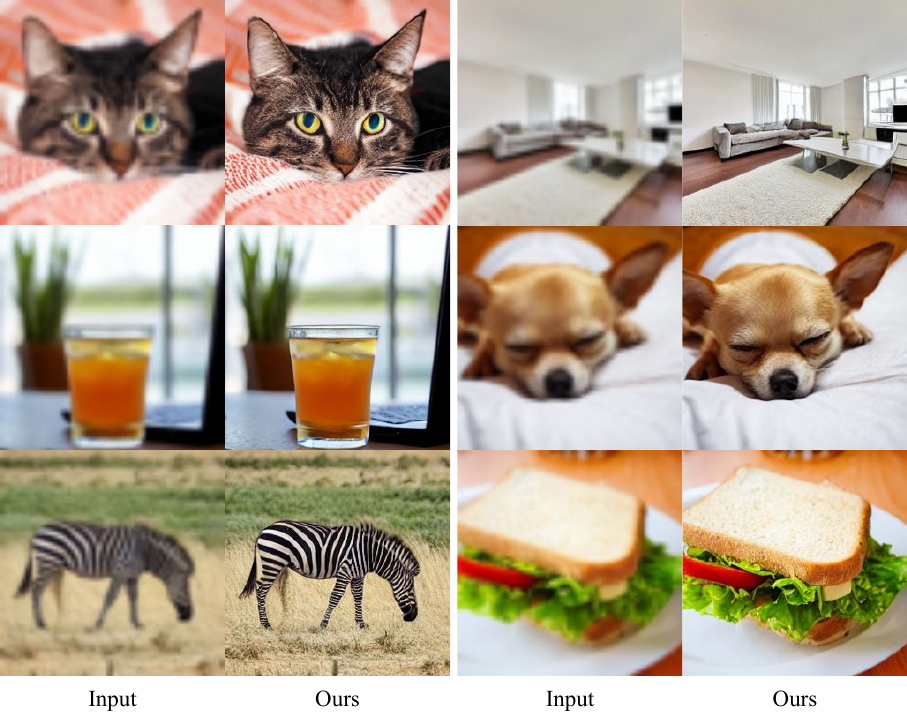}
    \vspace{-5mm}
    \caption{\textbf{Qualitative examples for image super-resolution.}
    }
    \label{fig:image_sr_qualitative}
\end{figure}

We provide qualitative examples for image super-resolution in Fig.~\ref{fig:image_sr_qualitative}.
Unlike traditional methods, which typically evolve from Gaussian noise and rely on concatenating upsampled low-resolution images as conditioning, our approach takes a more direct route: 
we demonstrate that it is possible to evolve a low-resolution image directly into a high-resolution image, eliminating the need for additional concatenation conditioning.

\label{sec:supp:sr}

\subsection{Ablation Study}

\begin{table}[]
\centering
\resizebox{0.9\columnwidth}{!}{
\begin{tabular}{l|c}
\bf Text encoder & \bf Recon. accuracy (\%) \\
\midrule
Text Encoder ($1\times1024$) & 95.12 \\
Text \VE ($1\times1024$) & 94.53 \\
\end{tabular}
}
\caption{\textbf{Ablation on text compression.}
Both text encoder and Text \VE preserve most of the input information, despite the large compression ratio ($77\times768 \rightarrow 1\times1024 $, $14.4\times$).
}
\label{tab:text_compression}
\end{table}

\label{sec:supp:abl}
\noindent\textbf{Text compression.}
In this section, we show that we can compress the input text embedding $x\in\mathbb{R}^{n\times d}$ into $z_0\in\mathbb{R}^{h \times w \times c}$ (\eg, $77\times768$ \CLIP tokens to $4\times32\times32$ latents for $256$px generation, $14.4\times$ compression) with a standard encoder or the proposed \VE while preserve most of the input information.
We report the per-token reconstruction accuracy, computed by cosine similarity, in Tab.~\ref{tab:text_compression}. 
The results show that both methods are effective at preserving the input information, achieving high reconstruction accuracy despite a large compression ratio.

\begin{figure*}[]
    \centering
    \includegraphics[width=\linewidth]{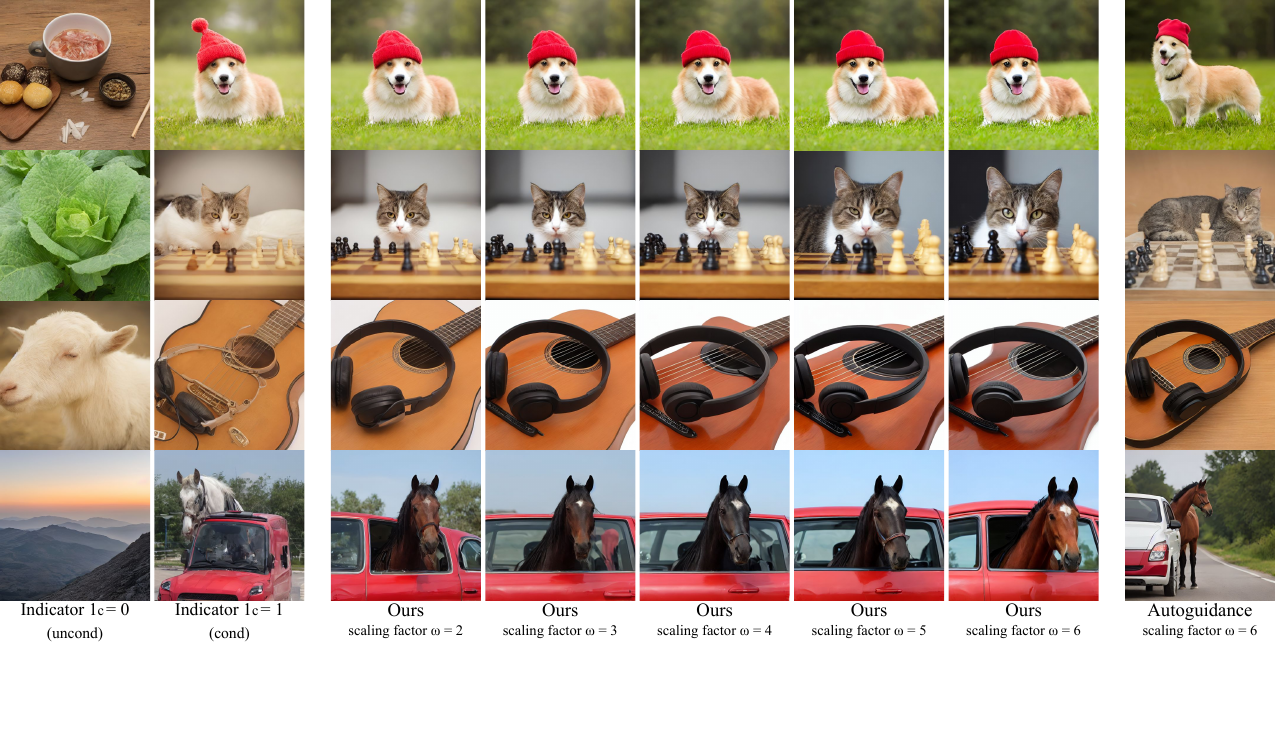}
    \vspace{-20mm}
    \caption{\textbf{Ablation on CFG with indicator.}
    The first two columns show the images generated when the indicator $1_c = 0$ (for unconditional generation) and $1_c = 1$ (for conditional generation), demonstrating that \OURS can still perform unconditional generation with the help of the indicator, thereby allowing for the use of standard CFG.
    We then demonstrate the improvement provided by CFG (middle five columns) and compare it with Autoguidance (last two columns).
    Prompts used to generate the images: \textit{`a corgi wearing a red hat in the park',`a cat playing chess',`a pair of headphones on a guitar',`a horse in a red car'}
    }
    \label{fig:supp_cfg}
\end{figure*}

\noindent\textbf{CFG indicator.}
In Fig.~\ref{fig:supp_cfg}, we study the effect of our CFG with indicator, and then compare our approach with Autoguidance~\cite{karras2024guiding}. 
The left two columns show the images generated when the indicator $1_c = 0$ (for unconditional generation) and $1_c = 1$ (for conditional generation).
It shows that despite generating an image by directly evolving from the text space into the image space without explicit conditioning, our model can still perform unconditional generation with the help of the indicator. This allows our model to support standard CFG.
Then, in the middle five columns, we show the images generated with different CFG scaling factors.
Similar to the standard \FM model, the CFG can significantly improve the image quality.
Finally, in the last two columns, we compare our CFG with indicator to Autoguidance, using the same scaling factor. 
Like our approach, Autoguidance also enables low-temperature sampling for models without explicit conditioning.
We observe that our CFG with indicator produces higher-fidelity images compared to Autoguidance.

\section{Additional Qualitative Examples}
\label{sec:supp:addl_qual}
\begin{figure*}[]
    \centering
    \includegraphics[width=0.87\linewidth]{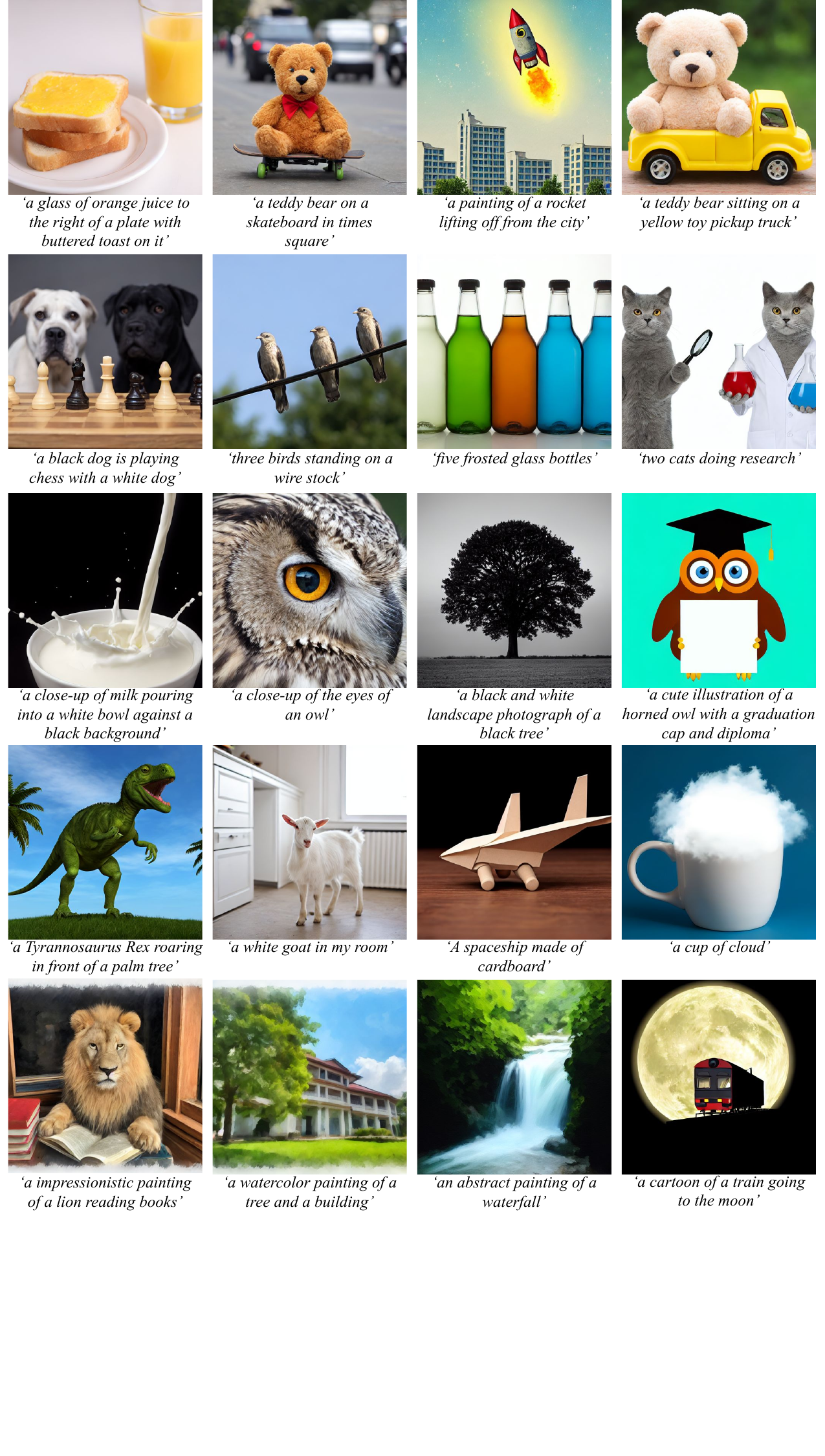}
    \vspace{-47mm}
    \caption{\textbf{Qualitative examples for \ttoi with \OURS.}
    }
    \label{fig:supp_t2i}
\end{figure*}

\begin{figure*}[]
    \centering
    \vspace{-5mm}
    \includegraphics[width=0.86\linewidth]{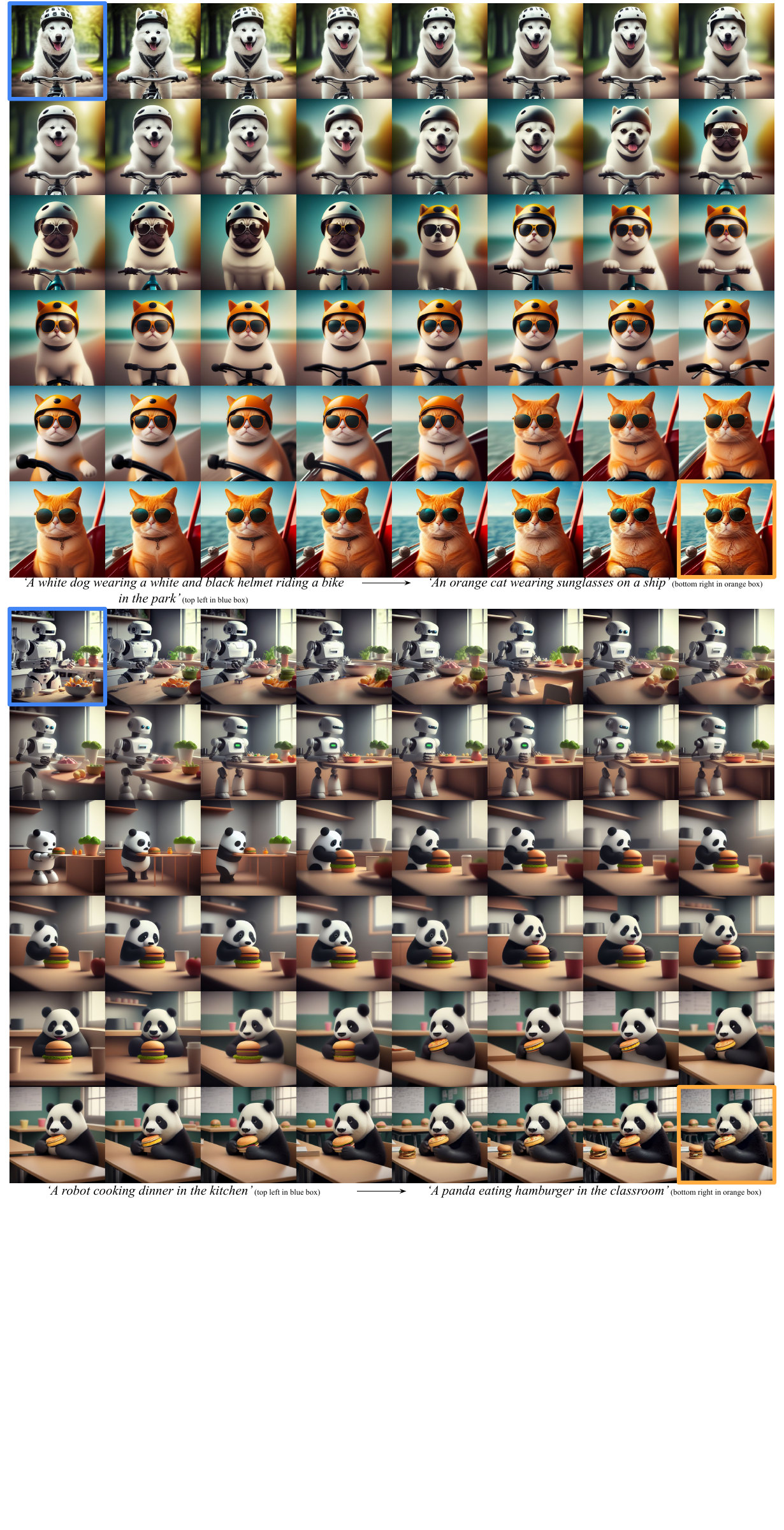}
    \vspace{-67mm}
    \caption{\textbf{Linear interpolation in latent space.}
    We show images generated by linear interpolation between two text latents (\ie, interpolation between $z_0$).
    Images generated by the first and second text latents are provided in the top-left and bottom-right corners.
    }
    \label{fig:supp_interp_2}
\end{figure*}

\begin{figure*}[]
    \centering
    \vspace{-5mm}
    \includegraphics[width=0.86\linewidth]{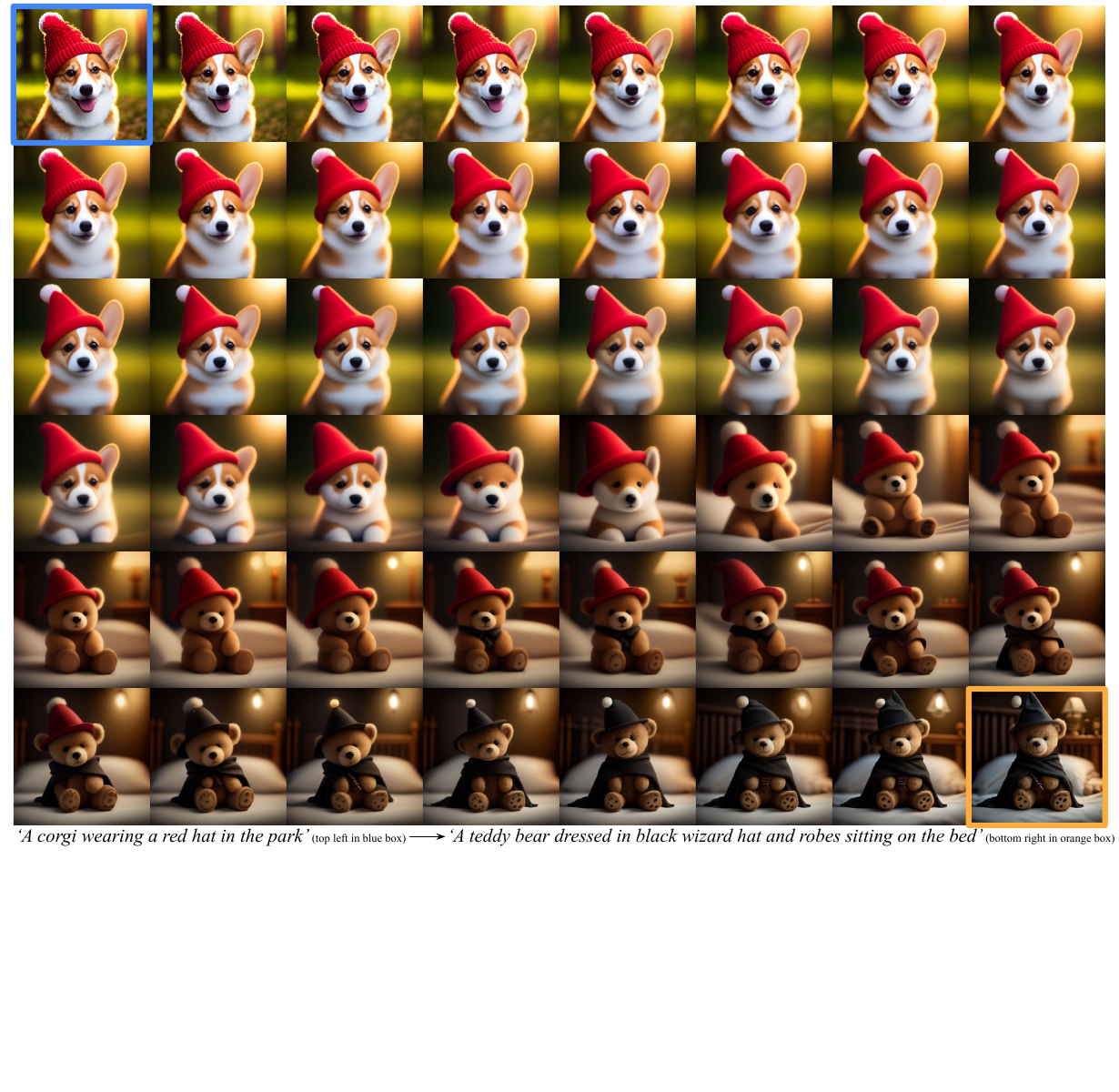}
    \vspace{-35mm}
    \caption{\textbf{Linear interpolation in latent space.}
    We show images generated by linear interpolation between two text latents (\ie, interpolation between $z_0$).
    Images generated by the first and second text latents are provided in the top-left and bottom-right corners.
    }
    \label{fig:supp_interp_3}
\end{figure*}

\begin{figure*}[]
    \centering
    \includegraphics[width=0.74\linewidth]{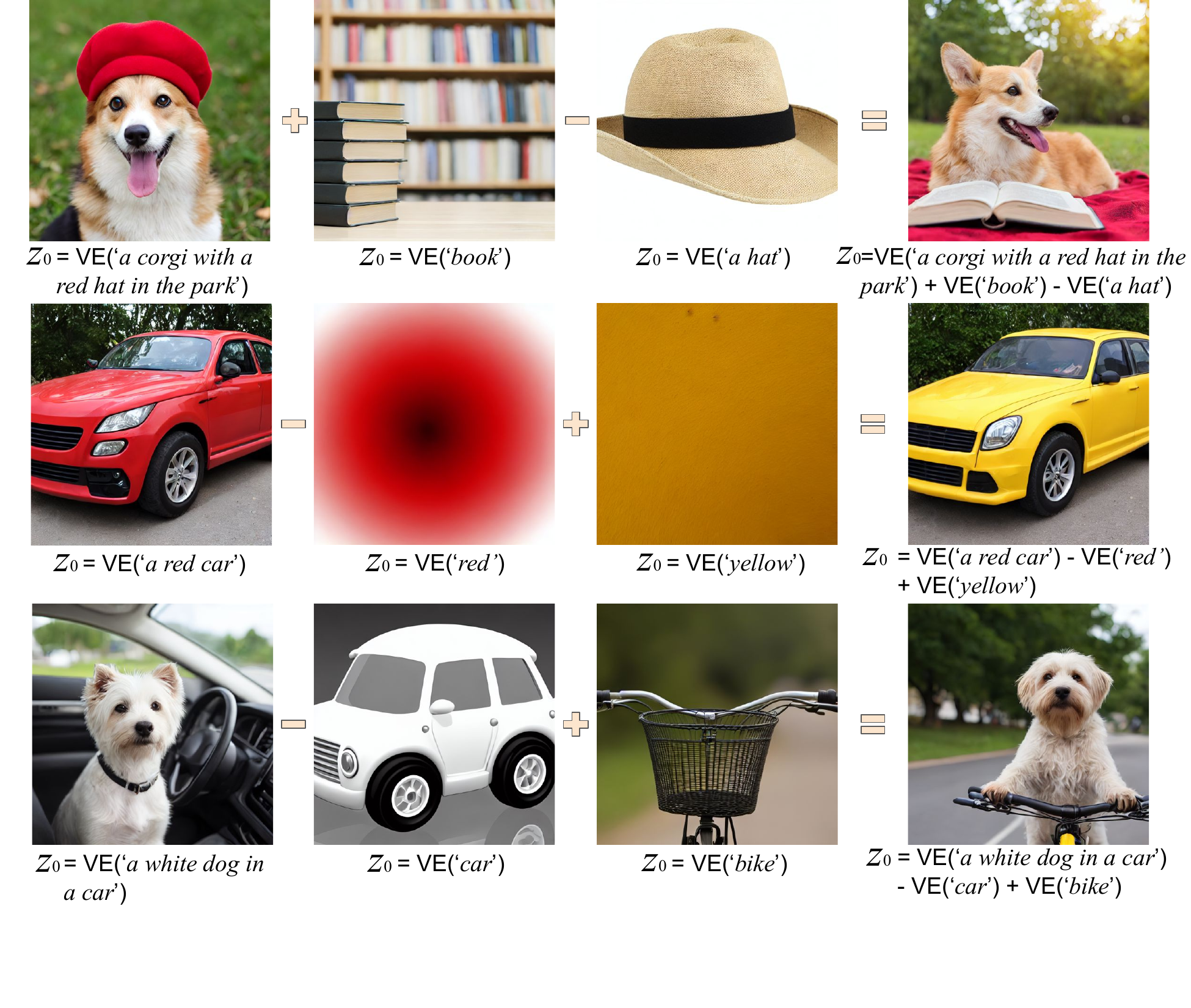}
    \vspace{-13mm}
    \caption{\textbf{Arithmetic in text latent space.}
    We map the text into the text latent space, perform arithmetic operations to obtain new latent representation, and use the resulting representation to generate the image.
    Latent $z_0$ used to generate each image is provided at the bottom.
    }
    \vspace{-3mm}
    \label{fig:supp_arithmetic}
\end{figure*}

We provide additional qualitative examples for \ttoi generation here. 
Specifically, we first provide $512\times 512$ images generated by our \OURS in Fig.~\ref{fig:supp_t2i}.
Next, we provide more examples for linear interpolation in latent space (Fig.~\ref{fig:supp_interp_2} and Fig.~\ref{fig:supp_interp_3}) and arithmetic operation in latent space (Fig.~\ref{fig:supp_arithmetic}).

\end{document}